\pdfoutput=1

\documentclass[11pt]{article}

\usepackage[final]{acl}
\usepackage{makecell}
\usepackage{times}
\usepackage{latexsym}

\usepackage[T1]{fontenc}

\usepackage[utf8]{inputenc}

\usepackage{microtype}

\usepackage{inconsolata}

\usepackage{graphicx}
\usepackage{times}
\usepackage{latexsym}
\usepackage{times}
\usepackage{latexsym}
\usepackage{booktabs}
\usepackage{boxedminipage}
\usepackage{amssymb}
\usepackage{amsmath}
\usepackage{todonotes} 
\usepackage{graphicx}
\usepackage{booktabs}
\usepackage{siunitx}

\usepackage{hyperref}
\usepackage{inconsolata}
\usepackage{pifont}%

\usepackage{tabu}
\usepackage{needspace}
\usepackage{fontawesome}
\usepackage{amsmath}
\usepackage{amsfonts}
\usepackage{amssymb}
\usepackage{enumitem}
\usepackage{listings}
\usepackage{xstring}
\usepackage{graphicx}
\usepackage{pbox}
\usepackage{subcaption}
\usepackage{epstopdf}
\usepackage{xstring}
\usepackage{multirow}

\usepackage{soul}
\usepackage{color}
\usepackage{graphicx}
\usepackage{multirow}
\usepackage{comment}
\usepackage{todonotes}
\usepackage{listings} %
\usepackage{graphicx}
\usepackage{subcaption}
\usepackage{verbatim}

\lstdefinestyle{smalllisting}{
    basicstyle=\small\ttfamily
}

\usepackage{xcolor}
\usepackage{listings}

\lstdefinestyle{aclprompt}{
  basicstyle=\ttfamily\small,
  columns=fullflexible,
  breaklines=true,
  breakatwhitespace=true,
  showstringspaces=false,
  keepspaces=true,
  frame=single,
  framerule=0.4pt,
  rulecolor=\color{black!25}
}
\usepackage{xcolor} %
\definecolor{lightblue}{rgb}{.50,.90,0.51}
\definecolor{tri}{rgb}{.25,.88,.82}
\definecolor{lilac}{rgb}{0.85,0.64,0.85}
\definecolor{atomictangerine}{rgb}{1.0, 0.6, 0.4}

\newcommand{\memel}{\emph{MemeLens}}

\usepackage{makecell}
\title{MemeLens: Multilingual Multitask VLMs for Memes}

\author{
Ali Ezzat Shahroor$^1$\thanks{~Equal contribution.},
Mohamed Bayan Kmainasi$^1$$^*$,
Abul Hasnat$^2$$^,$$^3$, \\
{\bf Dimitar Dimitrov$^4$, Giovanni Da San Martino$^5$,
Preslav Nakov$^6$, Firoj Alam$^1$}\\[2pt]
$^1$Qatar Computing Research Institute, Hamad bin Khalifa University, Qatar\\
$^2$Blackbird.AI, USA \quad
$^3$APAVI.AI, France \quad
$^4$Sofia University ``St. Kliment Ohridski'', Bulgaria\\
$^5$University of Padova, Italy \quad
$^6$Mohamed bin Zayed University of Artificial Intelligence\\
\texttt{\{alsh34060, mkmainasi, fialam\}@hbku.edu.qa}\\
\texttt{mhasnat@gmail.com, ilijanovd@fmi.uni-sofia.bg}\\
\texttt{giovanni.dasanmartino@unipd.it, preslav.nakov@mbzuai.ac.ae}\\
{\small\href{https://huggingface.co/collections/QCRI/media-integrity-intelligence}
{https://huggingface.co/collections/QCRI/media-integrity-intelligence}}
}

\begin{document}
\maketitle
\begin{abstract}

Memes are a dominant medium for online communication and manipulation because meaning emerges from interactions between embedded text, imagery, and cultural context. Existing meme research is distributed across tasks (e.g., \textit{hate, misogyny, propaganda, sentiment, humour}) and languages, which limits cross-domain generalization. To address this gap, we propose \textsc{MemeLens}, a unified multilingual, multitask explanation-enhanced Vision-Language Model (VLM) for meme understanding. We consolidate $38$ public meme datasets, filter and map dataset-specific labels into a shared taxonomy of $20$ tasks spanning harm, targets, figurative/pragmatic intent, and affect. We present a comprehensive empirical analysis across modeling paradigms, task categories, and datasets. 
Our findings suggest that robust meme understanding requires multimodal training, varies substantially across semantic categories, and remains sensitive to over-specialization when models are fine-tuned on individual datasets rather than trained in a unified setting. 
We make the experimental resources\footnote{\href{https://github.com/MohamedBayan/MemeLens}{Code}~|~\href{https://huggingface.co/QCRI/MemeLens-VLM}{MemeLens-VLM}~|~\href{https://huggingface.co/datasets/QCRI/MemeLens}{MemeLens-Datasets}} publicly available to the community.

\end{abstract}

\section{Introduction}
\label{sec:introduction}

Memes are among the most widely shared forms of online content \cite{brody2023memesharing,Barnes2024}. By pairing an image with a small amount of overlaid text, memes communicate stance, sarcasm, in-group identity, or hostility. Their interpretation is compositional: meaning emerges from the \emph{interaction} between the textual overlay and the visual scene, often mediated by background knowledge and cultural context, challenging purely textual or visual approaches.

This has motivated a growing body of work on multimodal meme understanding, spanning hate and abuse \citep{kiela2020hatefulmemes}, misogyny \citep{fersini-etal-2022-semeval}, affect and humor \citep{sharma-etal-2020-semeval}, \emph{figurative language} \citep{xu-2022-metmeme}, and persuasion techniques \citep{dimitrov-etal-2024-semeval,alam-etal-2024-armeme} across diverse datasets, tasks, and languages in practice.

Alongside these advances, the current state of the art remains largely task- and language-specific. Most efforts are organized around \emph{a single dataset} and \emph{a single phenomenon} (often via one shared task), which introduces at least three practical barriers: \textit{(i)} limited transfer across label spaces, languages, and domains; \textit{(ii)} evaluation protocols that are difficult to compare or reproduce across benchmarks; and \textit{(iii)} missed opportunities for parameter sharing, where capabilities learned for one aspect of meme understanding (e.g., affect) could systematically benefit others (e.g., harassment detection).

\begin{figure*}[t]
    \centering
    \includegraphics[width=0.9\textwidth]{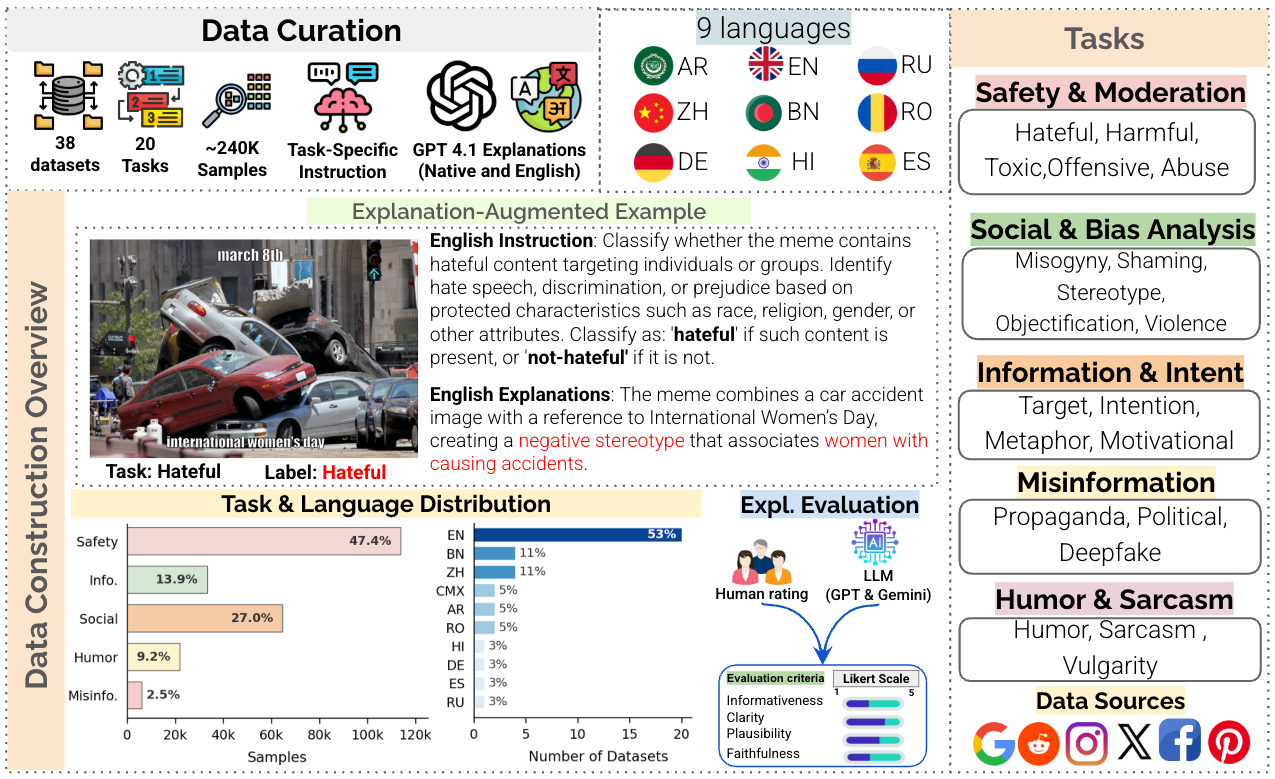}
    \vspace{-0.3cm}
\caption{\textbf{Overview of tasks and datasets in \textsc{MemeLens}.} Dataset-specific labels are mapped into a unified taxonomy for multi-task training and cross-dataset evaluation. Task categories include Safety, Info., Social, Humor, and Misinfo. Language abbreviations denote EN, BN, ZH, CMX (Hindi-English), AR, RO, HI, DE, ES, and RU.}    
    \label{fig:vllm_exp_meme}
    \vspace{-0.4cm}
\end{figure*}

In deployed settings (e.g., content moderation, real-time trend monitoring, and media literacy), models must \emph{simultaneously} reason about affect (sentiment and humor), pragmatic intent (e.g., sarcasm), and a broad spectrum of harm-related categories, while remaining robust to cultural and linguistic variation across platforms \citep{hee2024hsmoderation}.

Existing resources for meme understanding remain uneven across languages. As shown in Figure~\ref{fig:vllm_exp_meme}, most datasets are English-centric, while non-English benchmarks are much more limited. Cross-lingual generalization is further hindered by mismatches in label definitions, annotation granularity, and guidelines across datasets \citep{bui-etal-2025-multi3hate}. As a result, naive dataset mixing can be unreliable 
and may cause negative transfer 
in multilingual multitask training. At the same time, prior work argues that deeper meme understanding requires going beyond %
lexical and visual cues to infer latent intent and implied meaning \citep{ijcai2023p665}.

\needspace{5\baselineskip}

Motivated by these gaps, we frame meme understanding as a problem of \emph{multilingual, multimodal, multitask learning}, where models learn shared representations while harmonizing heterogeneous taxonomies across datasets to enable robust generalization across domains and settings, improving transfer, consistency, and performance. Concretely, we compile an extensive collection of meme datasets (38 sources in our current collection), spanning multiple languages and tasks.

This unification also raises an empirical question: \emph{which} modeling paradigm best supports multilingual multitask meme understanding under realistic heterogeneity?
To address it, we benchmark \textit{(i)} unimodal baselines (text-only, image-only), \textit{(ii)} multimodal fusion and sequence-based architectures, and \textit{(iii)} causal/instruction-style VLMs, analyzing not only aggregate accuracy but also cross-task and cross-lingual transfer.
\textit{Finally}, we situate our study relative to emerging efforts that emphasize robustness and generalization in meme understanding \citep{liu-etal-2025-mind,chen-etal-2025-adammeme}.
We summarize our main contributions below:
\begin{itemize}[noitemsep,topsep=0pt,leftmargin=*,labelsep=.5em]
    \item We curate and consolidate 38 publicly available meme datasets, applying rigorous filtering and a unified annotation mapping to create a coherent, multilingual, multi-task resource. 
    \item We introduce a unified taxonomy spanning 9 languages and 20 tasks, with consistent training and evaluation for cross-dataset benchmarking.
    \item We formulate meme understanding as \textbf{multilingual, multimodal and multitask learning} and present a unified training setup over a large dataset mixture curated under a consistent \emph{text-over-image} meme definition.
    \item We provide explanation-enhanced supervision and a systematic study across modeling paradigms, including unimodal baselines, multimodal sequence models, and large VLMs, evaluated under a unified training and benchmarking framework.
    
\end{itemize}
\paragraph{Our empirical analysis yields several key observations.}
We find that \textit{robust meme understanding consistently benefits from multimodal training} but varies substantially across semantic task categories and datasets. Task families involving implicit or rhetorical meaning, such as humor and sarcasm, remain challenging across all paradigms. Additionally, we observe that models fine-tuned on individual datasets tend to over-specialize, \textit{motivating unified multitask training when broad coverage and cross-dataset robustness are required}.

\section{Related Work}
\label{sec:related_work}

\noindent\textbf{Meme benchmarks.}
A major driver of progress in computational meme understanding has been curated datasets and shared tasks that frame meme semantics as supervised prediction~\cite{ijcai2022p781}. The \citet{kiela2020hatefulmemes} Hateful Memes Challenge showed the limitations of unimodal models, motivating approaches that integrate visual evidence with overlaid text. Beyond hate, SemEval expanded meme understanding to affective and pragmatic phenomena. Memotion \citep{sharma-etal-2020-semeval} targets sentiment and emotion (e.g., humor, sarcasm, offensiveness, motivation), while MAMI \citep{fersini-etal-2022-semeval} focuses on misogyny and fine-grained traits. SemEval also formalized persuasion and propaganda in Task 6 (2021) \citep{dimitrov-etal-2021-semeval} and later extended it to multilingual settings in Task 4 (2024) \citep{dimitrov-etal-2024-semeval}. 

\noindent\textbf{Multilingual and multicultural meme understanding.}
While early meme benchmarks centered on English, subsequent work has emphasized the multilingual nature of meme culture and the limits of monolingual generalization~\cite{alam2022survey}. For instance, MUTE \citep{hossain-etal-2022-mute} targets Bengali and code-mixed hateful memes, and Multi$^3$Hate \citep{bui-etal-2025-multi3hate} offers a parallel multilingual meme dataset to study cross-cultural annotation differences and cross-lingual VLM behavior. 

\noindent\textbf{Multitask meme modeling.}
A smaller body of work studies \textit{multi-task meme understanding}. Early work by \citet{chauhan-etal-2020-one} proposed a multimodal multi-task architecture spanning humor, sarcasm, offensiveness, motivational content, and sentiment. More recent efforts introduce datasets and models that jointly model multiple facets within a unified setting (e.g., hate, targets, stance, and humor) \citep{shah2024memeclip}, and benchmarks that evaluate VLMs across diverse meme tasks (e.g., humor and sarcasm) \citep{gavit2025vlms}. Despite these advances, most studies still train and report results \textit{separately per dataset and per phenomenon}.

\noindent\textbf{Explanations, roles, and reasoning.}
Meme semantics is often implicit and context-dependent, motivating approaches beyond surface classification, including connotation modeling and explanation generation \citep{pandiani-etal-2025-toxicmemes}. Prior work also shows that harmfulness can be satirical and highly context-sensitive \citep{pramanick-etal-2021-detecting}.
Several datasets use explanations as supervision, including hateful reasons for conditional rationale generation \citep{ijcai2023p665} and explanations for entity and role understanding \citep{sharma-etal-2023-aaai-exclaim}. More recent resources elicit richer meaning representations using multimodal QA \citep{agarwal-etal-2024-mememqa}, interpretation-augmented captioning \citep{park-etal-2025-memeinterpret}, and bilingual detection-plus-explanation setups \citep{kmainasi-etal-2025-memeintel,gu-etal-2025-mememind}. In parallel, model-centric work distills or structures reasoning traces to improve prediction and interpretability \citep{lin-etal-2023-beneath,lin-etal-2024-explainhm,hee-lee-2025-intmeme,bayan2026can,kmainasi2026adaptingreinforcementlearningchainofthought}. At the same time, explanations are not always helpful, as rationales can drift from meme evidence and \textit{explain-then-detect} pipelines may underperform without reasoning-aware objectives \citep{lu-etal-2025-rationales,mei-etal-2025-expohm}.

\noindent\textbf{VLMs and meme-specific adaptation.}
With the rise of strong VLMs, meme research has increasingly focused on adapting and prompting these models. Prompt-based approaches leverage the implicit knowledge of language models for hateful meme detection \citep{cao-etal-2022-prompting}.
Retrieval-based methods target out-of-domain generalization in evolving meme ecosystems \citep{mei-etal-2024-rgcl,mei-etal-2025-rahmd}. Agentic and multi-agent paradigms explore zero-shot or adaptive evaluation for harmful memes \citep{liu-etal-2025-mind,chen-etal-2025-adammeme}. Finally, meme-centered safety benchmarks for VLMs improve ecological validity by using real memes to probe harmful outputs \citep{lee-etal-2025-memesafetybench}.

\section{Dataset}
\label{sec:dataset}

\subsection{Dataset Curation}
\noindent
\paragraph{Language and task coverage.}
We curate a multilingual, multimodal collection of 38 publicly available meme datasets spanning nine languages such as English, Arabic, Bengali, Chinese, German, Spanish, Hindi, Romanian, and Russian, including both monolingual and code-mixed settings. These datasets cover 20  tasks as shown in Figure \ref{fig:vllm_exp_meme}.

\paragraph{Filtering.}
The source datasets vary substantially in annotation schemes, label schema, and even in what is considered a ``meme.'' Following prior work, we operationalize memes as images with embedded/overlaid text (image-text pairs) \citep{fersini-etal-2022-semeval,sharma-etal-2020-semeval,kiela2020hatefulmemes}. Accordingly, we formulate all tasks as \emph{multimodal} and require an explicit text modality (either released by the dataset or obtained via OCR) to ensure consistent cross-dataset processing and reliable downstream modeling.
In our curated set, we observed that some datasets include a non-trivial fraction of images that do not conform to this definition (e.g., images without embedded text; MMHS~\cite{Gomez2020exploring}) and/or do not release extracted textual content (e.g., BanglaAbuseMeme~\cite{das-etal-2023-banglaabusememe}, MIMIC Islamophobia~\cite{islam2024mimic}). We therefore identify and remove samples with empty text by extracting text with a language-specific OCR pipeline based on EasyOCR.\footnote{\url{https://github.com/JaidedAI/EasyOCR}}
 We choose EasyOCR due to its public availability and its use in closely related prior work \citep{alam-etal-2024-armeme}. We perform filtering independently within each dataset split (train/validation/test).
Across datasets, we remove $\sim$92K samples with empty textual content, mostly from the MMHS dataset, which includes many images without embedded text. This filtering improves visual-textual alignment while preserving the original dataset splits for the remaining instances.

\paragraph{Label taxonomy.}
We design a taxonomy that \textit{(i)} spans the task inventory observed across meme datasets, \textit{(ii)} accommodates heterogeneous label structures (binary, multi-class, multi-label), and \textit{(iii)} enables \emph{semantic alignment} across datasets through explicit label definitions. The taxonomy serves as a \emph{supervision schema} rather than a claim that a fixed ontology can exhaustively capture meme meaning.

\paragraph{Unified taxonomy and label harmonisation.}
Our taxonomy aligns heterogeneous benchmarks by defining a shared task inventory with 20 tasks and by normalising dataset-specific label encodings and surface forms such as \texttt{0/1}, yes/no, naming variants, and dataset-specific subclasses into canonical labels. This reduces spurious variation introduced by annotation conventions and supports reliable multilingual multitask training and cross-benchmark comparison.

\paragraph{Task families.}
We group the 20 tasks into five semantic categories.
\begin{enumerate}[noitemsep,topsep=0pt,leftmargin=*]
    \item \textbf{Safety \& Moderation} hateful, harmful, toxic, offensive, abuse, vulgarity
    \item \textbf{Social \& Bias} misogyny, shaming, stereotype, objectification, violence
    \item \textbf{Information \& Intent} target, intention, metaphor, motivational
    \item \textbf{Misinformation} propaganda,  deepfake %
    \item \textbf{Humor \& Sarcasm} humor, sarcasm
\end{enumerate}

\paragraph{Multi-label normalisation.}
For multi-label settings like the misogyny categories in MIMIC, we normalise composites by decomposing them into shared primitives, for example \texttt{prejudice+humiliation} $\rightarrow$ \{\texttt{prejudice}, \texttt{humiliation}\}. We retain an \texttt{unspecified} label when the source annotation is explicitly underspecified. This preserves the original signal without forcing incompatible granularity into a single label space.

\paragraph{Mapping policy.}
Table~\ref{tab:label-mapping} summarises the harmonisation used to construct \textsc{MemeLens}. Since constituent datasets adopt heterogeneous schemes ranging from binary indicators such as \texttt{0/1} and yes/no to fine-grained taxonomies, we map each original label to a human-readable canonical form. When the original label space is reduced, marked with $\dagger$, we either collapse rare or ambiguous subclasses into a canonical task label, for example, MMHS 41-way $\rightarrow$ \texttt{hateful} and \texttt{not-hateful}, or remove under-represented residual categories such as \textit{other} in MET-Meme En and \textit{unspecified} in HarMeme. Canonical labels follow a consistent \texttt{positive} and \texttt{not-positive} convention to support unified evaluation across datasets and tasks.

\paragraph{Label cardinality and task formulation.}
Our taxonomy supports binary, multi-class, and multi-label supervision (reported in Table~\ref{tab:dataset_stats}), so we do not convert all tasks into binary form. When the original label space is semantically meaningful and stable, we preserve it. Examples include the 3-way \textsc{RoMemes}~\cite{puaics2024romemes} deepfake labels \texttt{real}, \texttt{fake}, and \texttt{deepfake}, the multi-class harmfulness levels and structured target categories in \textsc{HarMeme}~\cite{pramanick-etal-2021-detecting}, and the 7-way misogyny-category annotations in MIMIC. For multi-label datasets, we retain the signal using a standard one-vs-rest formulation implemented as per-label binary heads.

For the remaining tasks, we adopt a binary canonical form to support capability-oriented benchmarking across heterogeneous sources. Many datasets differ only in label encodings such as yes/no and \texttt{0/1}, or use incompatible granularities for the same phenomenon. Mapping to a shared binary form such as \texttt{hateful} and \texttt{not-hateful} removes mismatches and enables fair cross-dataset training and evaluation. In addition, multi-class schemes reflect dataset-specific intensity scales or annotation artifacts not consistently defined across benchmarks. Collapsing these to a coarse binary decision tests whether a model detects the core phenomenon rather than fitting a grading rubric.

We also normalize semantically equivalent label variants such as \textit{no\_harmful}, \textit{not\_harmful}, and \textit{non-hateful} into a single canonical label to reduce variation driven by naming conventions rather than underlying phenomena.

\subsection{Dataset Statistics}
After preprocessing and filtering, the benchmark comprises approximately 178K, 22K and 40K instances for train/validation/test, respectively. Most datasets use binary labelings, while a smaller subset uses multi-class schemes. The number of labels ranges from 2 to 7, depending on the task and original annotation design. 

\begin{figure}[t]
    \centering
    \includegraphics[width=0.98\columnwidth]{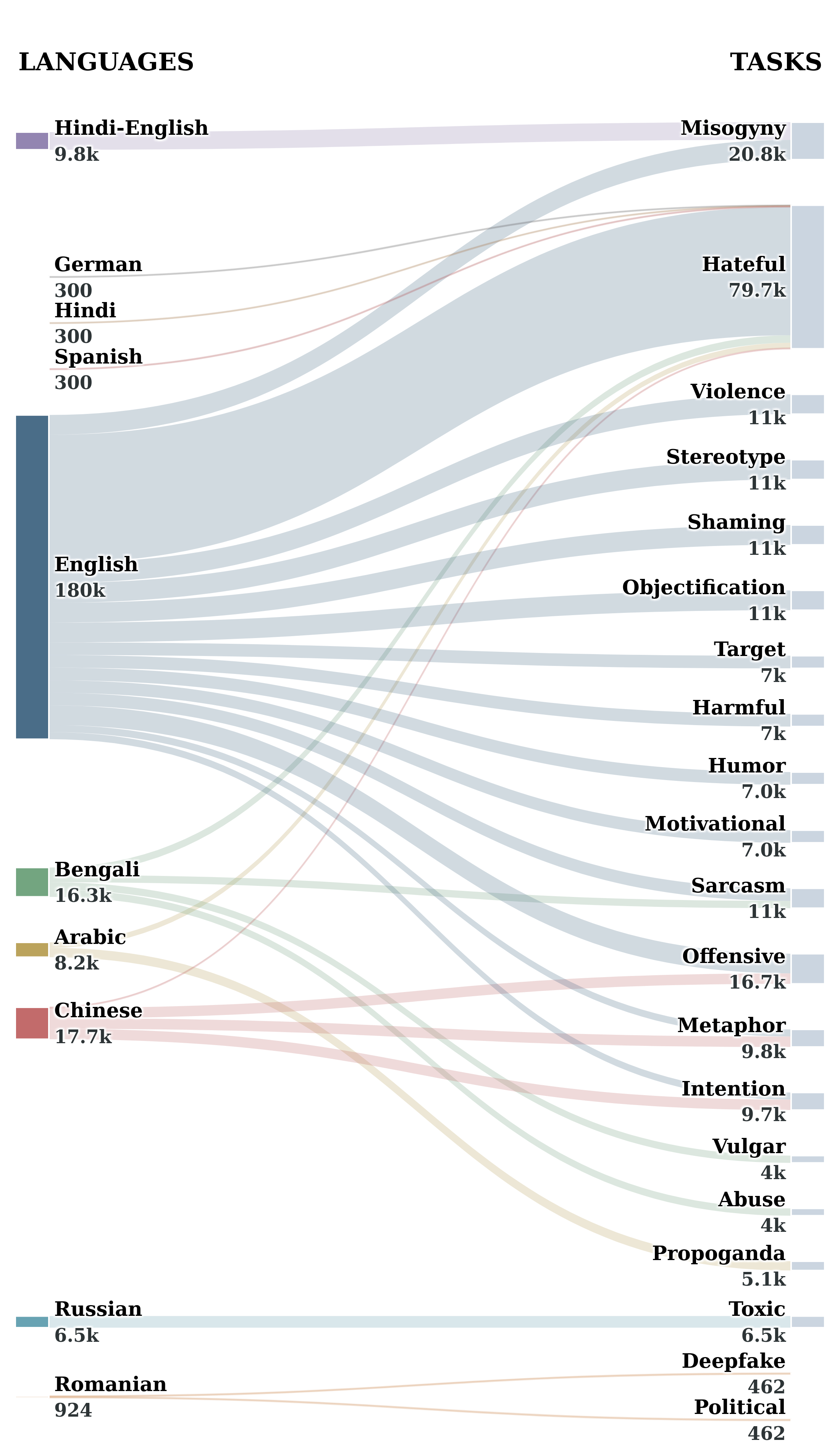}
    \vspace{-0.1cm}
    \caption{Language and task coverage in \textsc{MemeLens}. 
    }
    \label{fig:Language_task}
    \vspace{-0.3cm}
\end{figure}

Table~\ref{tab:dataset_stats} gives detailed dataset statistics. Figure~\ref{fig:Language_task} shows how instances are distributed across source languages and downstream tasks. The collection exhibits strong variation in language coverage. English accounts for the largest portion of instances at roughly 180K and spans most task categories, with a large portion in hate-related tasks at roughly 79K. Other languages are more task-focused. Bengali, Arabic, and Russian primarily map to offensive, misogyny, and toxic tasks, respectively, while Chinese is largely task-focused on propaganda, metaphor, and intention across datasets with varying annotation schemes and domain-specific characteristics. 
Hate- and harm-related tasks constitute a substantial portion of the benchmark across languages. The largest single source is \texttt{MMHS} with approximately 41K, 6K, and 12K instances in train, validation, and test, and several other datasets provide multilingual coverage, including \texttt{FHM}, \texttt{MIMIC Islamophobia}, \texttt{MUTE}, and \texttt{Multi3Hate}. The diversity in datasets and label structures supports analyses of cross-dataset transfer and taxonomy mismatch, while preserving the original annotation schemes aside from label-name normalization.

\subsection{Explanation Augmentation}
\paragraph{Explanation generation.} We additionally developed explanations for all 38 datasets in \memel{}. Each augmented instance pairs the original label with a short natural-language rationale/explanation that justifies the decision and explicitly grounds it in \emph{both} the visual content and the overlaid/extracted text, enabling evaluation under standard label prediction as well as explanation-aware settings.
Following \citet{kmainasi-etal-2025-memeintel}, we generate an explanation $e$ by sampling from the conditional model
$e \sim p_\theta\!\left(e \mid \mathbf{x}^{(I)},\, \mathbf{x}^{(T)},\, \mathbf{g}_t\right)$,
where $\mathbf{x}^{(I)}$ is the meme image, $\mathbf{x}^{(T)}$ is its text modality, and $\mathbf{g}_t$ denotes task-specific annotation guidelines.
To support multilingual and cross-cultural analysis, we provide explanations in English and, for non-English datasets, also in the original language, using task-specific prompts to ensure consistency across diverse phenomena (e.g., toxicity, misinformation, intent, and humor).

We generate explanations using GPT-4.1 \cite{openai2023gpt4}. The detailed prompts are provided in Appendix~\ref{sec_propmpts} (Tables~\ref{tab:app_meme_expl_user_prompt_en}--\ref{tab:app_meme_expl_prompt}). This choice is supported by \citet{kmainasi-etal-2025-memeintel}, who found that GPT-4.1 explanations can serve as high-quality references, with average human ratings above 4/5 for faithfulness, clarity, plausibility, and informativeness.

Overall, this augmentation extends \memel{} beyond label-only supervision and supports research on explainable multimodal learning, reasoning-centric evaluation, and explanation-conditioned training and inference. For explanation generation, we use zero-shot prompting with deterministic decoding by setting the temperature to $0$ to ensure reproducibility. Explanations are constrained to $\sim$114 words on average, with mean lengths of 118 words for English and 104 for native-language explanations. Detailed explanation statistics are reported in Appendix~\ref{app:explanation_stats}.

\paragraph{Explanation quality.}
We evaluate explanations using four criteria: \emph{informativeness}, \emph{clarity}, \emph{plausibility}, and \emph{faithfulness}, each scored on a 1-5 Likert scale (higher is better), following the guidelines of~\citet{kmainasi-etal-2025-memeintel,hasanain-etal-2025-propxplain}. As human rating at MemeLens scale (38 datasets) is prohibitively costly, we use a complementary automatic assessment with two strong LLM judges, \textbf{Gemini 2.5 Pro} \cite{comanici2025gemini} and \textbf{GPT-5} \cite{singh2025openai}, to reduce single-judge bias. 

In Table~\ref{tab:judge_scores}, we report the results. Both judges assign consistently high scores across all criteria, with GPT-5 achieving a higher overall mean than Gemini (4.65 vs.\ 4.36). The largest gaps appear in clarity (4.80 vs.\ 4.57) and faithfulness (4.69 vs.\ 4.30). The shared ranking across judges suggests the explanations are generally coherent, informative, and grounded in the multimodal input.

In Table~\ref{tab:human_metrics}, we report manual evaluation on a small sample of generated explanations from four representative datasets. Although the human scores are generally lower than the LLM-judge scores, they show a broadly consistent trend. FHM performs best, with strong and balanced scores across all criteria, while ArMeme is clearly more difficult, particularly in plausibility and faithfulness. Multi3 and MIMIC show moderate performance. Across datasets, clarity is typically higher than plausibility and faithfulness, suggesting that the model often produces explanations that are easy to follow but less reliably grounded. Overall, the human evaluation supports the broader picture from the automatic analysis while showing that explanation quality varies substantially by dataset.

\section{Experiments}
\label{sec:experiments}

\subsection{Instruction Dataset}
\label{subsec:instruction-dataset}

To construct instruction-following datasets, we begin with a single manually designed seed instruction for each dataset. We then expand this seed using two LLMs, GPT-4.1 and Gemini-3-Pro, each generating 10 paraphrased instructions, resulting in approximately 20 English instructions per dataset. 

We prepare two different output formats. The \textit{classification-only} variant asks the model to output only the \texttt{Label}. The \textit{classification-with-explanation} variant follows \citet{kmainasi-etal-2025-memeintel,hasanain-etal-2025-propxplain} and outputs \texttt{Label} and \texttt{Explanation} in a fixed structure. In both cases, we use a short system prompt that specifies the task and enforces the required format.

\subsection{Baselines}
\label{subsec:baselines}

\noindent\textbf{Unimodal baselines.}
For text-only modeling, we fine-tune \texttt{bert-base-multilingual}~\cite{devlin-etal-2019-bert} on OCR %
text. For image-only modeling, we use \texttt{ViT-B/16}~\cite{dosovitskiy2021vit,radford2021clip} trained solely on visual inputs.

\noindent\textbf{Multimodal baselines.}
We fine-tune \texttt{Qwen3-VL-8B-Instruct} (Qwen3-VL-8B)~\cite{bai2025qwen3} separately for each dataset using a dataset-specific sequence-classification (\texttt{seq\_cls}) head. Unlike \textsc{MemeLens}, this baseline does not share parameters across datasets nor perform unified multitask learning.

\noindent\textbf{Zero-shot MLLM baselines.}
We additionally evaluate instruction-tuned MLLMs in a zero-shot setting, 
including GPT-4.1 and Qwen3-8B.

\subsection{Training Setup}
\label{subsec:training-setup}

\noindent\textbf{Unimodal models.}
We fine-tune text-only and image-only baselines with standard supervised training. We use batch size 32, learning rate $3\times10^{-5}$, and weight decay 0.01, and train for 7 epochs. We tune hyperparameters on the development set and select the checkpoint with the best development performance for test evaluation.

\noindent\textbf{Multimodal models.}
We fine-tune all multimodal models with LoRA \citep{hu2021loralowrankadaptationlarge} and use a \textit{two-stage training} following \citet{kmainasi-etal-2025-memeintel}. \textbf{\textit{Stage~I}} (classification) optimizes label prediction only. \textbf{\textit{Stage~II}} continues from the best Stage~I checkpoint and jointly optimizes label prediction and explanation generation, where the explanation loss acts as an auxiliary regulariser while keeping the inference conditioning for classification unchanged. More details on the experimental parameters are provided in Appendix~\ref{sec_app_exp_details}.

For the \texttt{seq\_cls} baseline, we use the same setup, but train a dataset-specific head adapter for 20 epochs with learning rate $1\times10^{-5}$.

\section{Results and Discussion}
\label{sec:results}

In this section we report results for \memel{} across multiple modeling paradigms and analyze \textit{(i)} the impact of multimodality, \textit{(ii)} variation across task families, \textit{(iii)} dataset-level differences relative to prior work, and transfer under unified and single-dataset fine-tuning across languages, domains, and varying levels of task complexity and annotation noise. We report Accuracy (Acc), Macro-F1 (M-F1) as the primary metric under class imbalance, and Weighted-F1 (W-F1).

\subsection{Overall Model Performance}
\label{subsec:overall_results}

In Table~\ref{tab:overall_results}, we compare uni-modal, multimodal, zero-shot, and fine-tuned models including \memel{}. Uni-modal text and image models achieve comparable performance, suggesting that each modality provides useful signals independently. However, the multimodal sequence-classification baseline consistently outperforms both uni-modal variants, highlighting the importance of cross-modal interaction for meme understanding in complex, context-dependent scenarios across diverse datasets and real-world applications.

\begin{table}[!tbh]
\centering
\setlength{\tabcolsep}{3pt} 
\scalebox{0.78}{%
\begin{tabular}{lccc}
\hline
\textbf{Model / Modality} & \textbf{Acc} & \textbf{M-F1} & \textbf{W-F1} \\
\hline
Uni-modal (Text) & 65.0 & 0.460 & 0.590 \\
Uni-modal (Image) & 62.7 & 0.454 & 0.592 \\
Multi-modal (Seq-Classification) & 70.6 & 0.578 & 0.679 \\
\hline
\multicolumn{4}{c}{\textbf{Zero-Shot}} \\
\hline
GPT-4.1 & 61.2 & 0.533 & 0.599 \\
Qwen3-VL-8B-Instruct \cite{bai2025qwen3} & 55.1 & 0.482 & 0.539 \\
InternVL3.5-8B \cite{wang2025internvl3} & 55.4 & 0.476 & 0.545 \\
Gemma-3-12B \cite{team2025gemma} & 48.2 & 0.439 & 0.485 \\
Qwen3-2B \cite{xu2025qwen3} & 45.6 & 0.394 & 0.431 \\
Phi-3.5-Vision-4.2B \cite{abdin2024phi3technicalreporthighly} & 43.8 & 0.393 & 0.447 \\
\hline
\textsc{\textbf{MemeLens}} & \textbf{74.1} & \textbf{0.625} & \textbf{0.720} \\
\hline
\end{tabular}
}
\vspace{-0.2cm}
\caption{Competitive performance across models on the full \textbf{test set}.}
\label{tab:overall_results}
\vspace{-0.3cm}
\end{table}

Pretrained zero-shot models demonstrate competitive but inconsistent performance. GPT-4.1 performs similarly to the fine-tuned unimodal baselines despite no task adaptation; however, Qwen3-VL-8B lags behind, suggesting that instruction tuning alone does not match task-specific fine-tuning for meme understanding across datasets with varying complexity, label distributions, and multimodal reasoning requirements in practice.

Fine-tuning yields the largest gains. \textsc{\textbf{MemeLens}}, obtained by fine-tuning Qwen3-VL-8B on the unified multitask multimodal mixture, achieves the best overall results and produces label predictions with explanations for interpretability. 

\noindent\textbf{MemeLens vs.\ Gemma-4 (fine-tuned).}
To better understand whether a Qwen3-based model generalizes more effectively than another strong model, we also fine-tune Gemma-4 \cite{team2026gemma}. The fine-tuned model achieves an accuracy of 74.3\%, nearly identical to the 74.1\% obtained by MemeLens. The difference is not statistically significant according to the Wilcoxon signed-rank test.

\subsection{Semantic Task Analysis}
\label{subsec:taskwise}
For task-wise analysis, we group datasets into five high-level semantic categories (Figure~\ref{fig:vllm_exp_meme}): 
\textit{Safety \& Moderation} (e.g., hateful, harmful, toxic, offensive, abuse), 
\textit{Social \& Bias Analysis} (e.g., misogyny, shaming, stereotyping, objectification, violence), 
\textit{Information \& Intent} (e.g., target identification, intention, metaphor, motivational content), 
\textit{Misinformation} (e.g., propaganda, political manipulation, deepfakes), and 
\textit{Humor \& Sarcasm} (e.g., humor, sarcasm, vulgarity). 
This grouping reflects shared semantic properties across datasets.

\begin{table}[t]
\centering
\setlength{\tabcolsep}{3pt}
\renewcommand{\arraystretch}{1.05}
\scalebox{0.82}{%
\begin{tabular}{lccccc}
\hline
\textbf{Task Category} & \textbf{\small{Text}} & \textbf{\small{Image}} & \textbf{\small{MM-Seq}} & \textsc{\textbf{\small{MmLs}}} & \textbf{\small{Qzs}} \\
\hline
Safety \& Moderation      & 0.45 & 0.47 & 0.57 & \textbf{0.61} & 0.52 \\
Social \& Bias   & 0.50 & 0.57 & 0.69 & \textbf{0.77} & 0.61 \\
Information \& Intent     & 0.51 & 0.42 & 0.53 & \textbf{0.60} & 0.32  \\
Misinformation            & 0.43 & 0.49 & 0.60 & \textbf{0.67} & 0.57 \\
Humor \& Sarcasm          & 0.41 & 0.42 & 0.46 & \textbf{0.63} & 0.30 \\
\midrule
\textbf{Average} & 0.46 & 0.47 & 0.57 & \textbf{0.65} & 0.46 \\
\hline
\end{tabular}%
}
\vspace{-0.2cm}
\caption{Task-wise comparison across models on the full test set. \textsc{\textbf{MmLS}} refers to \textsc{\textbf{MemeLens}}, and \textbf{Qzs} denotes the non-fine-tuned (zero-shot) Qwen3-VL-8B-Instruct model. Scores are reported as \textbf{Macro-F1}.}
\label{tab:taskwise_results}
\vspace{-0.2cm}
\end{table}

\begin{table}[t]
\centering
\setlength{\tabcolsep}{4pt} %
\scalebox{0.82}{%
\begin{tabular}{lcccc}
\toprule
\textbf{Language} & \textbf{Text} & \textbf{Image} & \textbf{MM-Seq} & \textsc{\textbf{MemeLens}} \\
\midrule
English  & 0.430 & 0.438 & 0.517 & \textbf{0.560} \\
Arabic   & 0.533 & 0.490 & \textbf{0.666} & 0.613 \\
Chinese  & 0.539 & 0.385 & 0.618 & \textbf{0.664} \\
Hindi    & 0.579 & 0.528 & \textbf{0.750} & 0.724 \\
Spanish  & 0.358 & 0.661 & 0.600 & \textbf{0.796} \\
Bangla   & 0.592 & 0.635 & 0.706 & \textbf{0.723} \\
Romanian & 0.331 & 0.461 & 0.559 & \textbf{0.663} \\
German   & 0.371 & 0.504 & 0.714 & \textbf{0.731} \\
Russian  & 0.493 & 0.495 & \textbf{0.698} & 0.691 \\
\midrule
\textbf{Average} & 0.470 & 0.511 & 0.648 & \textbf{0.685} \\
\bottomrule
\end{tabular}
}
\vspace{-0.2cm}
\caption{Language-wise comparison across models. Scores are reported as \textbf{Macro-F1}.}
\label{tab:language_results}
\vspace{-0.4cm}
\end{table}

In Table~\ref{tab:taskwise_results}, we report Macro-F1 aggregated by task family. Multimodal models outperform unimodal baselines across all categories, and \textsc{MemeLens} achieves the best average performance. Comparing \textsc{MemeLens} with its unfine-tuned counterpart, Qzs, shows consistent gains across all categories. %

\textsc{MemeLens} improves most on \textit{Social \& Bias, Misinformation}, and \textit{Humor \& Sarcasm}, where visual context and implicit cues matter. \textit{Safety \& Moderation} also benefits from multimodality, while \textit{Information \& Intent} shows smaller but consistent gains. \textit{Humor \& Sarcasm} remains the most challenging, reflecting its reliance on pragmatic and cultural cues.

In Table~\ref{tab:language_results}, we report Macro-F1 by language. \textsc{MemeLens} gives the strongest average performance and typically improves over unimodal baselines, while seq\_cls remains competitive in several languages. Overall, unified multilingual training improves multimodal meme understanding, though transfer strength varies across languages, task categories, and model families.

\subsection{Dataset-Level SOTA Comparison}
\label{subsec:dataset_variability}
To assess robustness beyond aggregate metrics, we analyze performance at the dataset level and compare \textsc{\textbf{MemeLens}} against previously reported state-of-the-art (SOTA) results where available. As meme datasets adopt heterogeneous evaluation metrics, we group benchmarks according to their official evaluation metric (Acc, M-$F_1$, or $F_1$ on the positive class) and report performance under the corresponding metric.

For comparability, we exclude a small number of datasets for which prior work does not clearly specify the $F_1$ variant (e.g., Macro-$F_1$ vs.\ Weighted-$F_1$), and we additionally account for differences in dataset preprocessing. Our unified dataset removes samples without embedded text under a consistent text-over-image definition, which can affect comparability with results reported on unfiltered data.

We treat these comparisons as \emph{indicative} rather than strictly like-for-like, since our preprocessing and evaluated instances differ from those of some published studies. Under this caveat, \textsc{\textbf{MemeLens}} is broadly comparable to, and on several benchmarks exceeds, dataset-specific SOTA: aggregate gaps remain small under macro-$F_1$ ($\Delta \approx 0.01$) and $F_1$-POS ($\Delta \approx 0.02$), and on the accuracy-based benchmarks \textsc{\textbf{MemeLens}} exceeds the cited results on most datasets (per-dataset numbers in Appendix~\ref{appendix:accuracy_results}). These results suggest that unified multitask training can match strong specialized models on average despite heterogeneous tasks, labels, and languages.

At the same time, performance still varies across datasets, which is expected given differences in task formulation, label granularity, and domain cues. Overall, these findings highlight an inherent trade-off between unification and specialization. While dataset-specific models may achieve higher peak performance on individual benchmarks, a single unified model such as \textsc{\textbf{MemeLens}} provides strong overall performance and broad coverage across tasks and languages. Detailed dataset-level results are provided in Appendix~\ref{appendix:dataset_results}.

\subsection{Ablation: Single-Dataset Fine-Tuning}
\label{subsec:diagnostic_single_dataset}
To test whether single-dataset fine-tuning is sufficient for robust meme understanding, we fine-tune the Qwen3-8B model on the Facebook Hateful Memes (FHM) dataset \citet{kiela2020hatefulmemes} and compare it to \memel{}. This analysis assesses the extent to which training on a single dataset transfers across related tasks.
As shown in Table~\ref{tab:diagnostic_single_dataset}, single-dataset fine-tuning yields lower performance on \memel{} than unified multitask training. This suggests that models trained on a single benchmark tend to \textit{over-specialize} to dataset-specific distributions and annotation conventions, which limits transfer across tasks and languages even when the datasets share broad semantic overlap.

\begin{table}[t]
\centering
\setlength{\tabcolsep}{4pt} 
\scalebox{0.84}{%
\begin{tabular}{lccc}
\hline
\textbf{Training Setup} & \textbf{Acc} & \textbf{M-F1} & \textbf{W-F1} \\
\hline
\textsc{\textbf{MemeLens}} & \textbf{74.1} & \textbf{0.625} & \textbf{0.720} \\
FHM-only fine-tuning & 56.9 & 0.495 & 0.556 \\
\hline
\end{tabular}
}
\vspace{-0.2cm}
\caption{Comparison between unified multitask training and single-dataset fine-tuning. M-F1: Macro-F1, W-F1: Weighted-F1.}
\label{tab:diagnostic_single_dataset}
\vspace{-0.4cm}
\end{table}

Importantly, we present this experiment as an analysis of when unified training across heterogeneous tasks and datasets is beneficial. The findings provide empirical support for this direction while also revealing remaining challenges in generalization. Methods that explicitly improve cross-dataset and label-set transfer remain a promising direction for future work.

\subsection{Task and Language Generalization}
\label{subsec:loto_lolo}

The single-dataset analysis showed limited cross-dataset transfer. We therefore test whether unified training generalizes to unseen tasks and languages using two leave-one-out settings. Leave-One-Task-Out (LOTO) removes an entire task and its datasets from training, while Leave-One-Language-Out (LOLO) removes all data from one language. Models are evaluated only on the held-out task or language. We compare zero-shot Qwen3-VL-8B (ZS), the corresponding LOTO/LOLO model, and the full \memel{} model, using the same test set and metrics.

\begin{table}[!tbh]
\centering
\setlength{\tabcolsep}{4pt}
\scalebox{0.84}{%
\begin{tabular}{lcc}
\hline
\textbf{Training Setup} & \textbf{Acc} & \textbf{M-F1} \\
\hline
\textbf{Zero-shot} (Qwen3-VL-8B) & 53.8 & 0.471 \\
\quad + \textbf{FT} unseen task (LOTO) & 62.6 & 0.513 \\
\quad + \textbf{FT} unseen language (LOLO) & 62.3 & 0.518 \\
\textsc{\textbf{MemeLens}} (full) & \textbf{74.0} & \textbf{0.619} \\
\hline
\end{tabular}
}
\vspace{-0.2cm}
\caption{Generalization performance on held-out tasks and languages, averaged across 38 datasets using matched test instances.}
\label{tab:loto_lolo}
\vspace{-0.3cm}
\end{table}

As shown in Table~\ref{tab:loto_lolo}, both held-out settings substantially outperform the zero-shot model despite having no training data from the target task or language, recovering about $42$--$44\%$ of the accuracy gain obtained by full \memel{} training. The accuracy improvement over zero-shot is significant for both settings (one-sided Wilcoxon signed-rank; LOTO $p=0.001$, LOLO $p<10^{-4}$). In Macro-F1 the gain is significant for LOLO ($p=0.005$) but only marginal for LOTO ($p=0.052$), indicating that some task-specific or rare labels remain more difficult to generalize.

\begin{figure}[!tbh]
\centering
\includegraphics[width=\columnwidth]{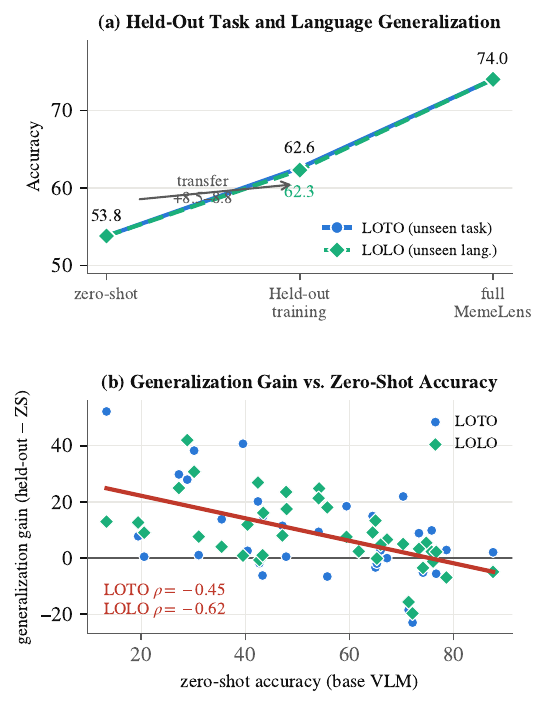}
\vspace{-0.3cm}
\caption{Generalization to held-out tasks and languages. \textbf{(a)} Performance of zero-shot, held-out, and full \memel{} models. \textbf{(b)} Generalization gain versus zero-shot performance.}
\label{fig:loto_lolo_transfer}
\end{figure}

Figure~\ref{fig:loto_lolo_transfer}(a) shows the overall trend: both LOTO and LOLO outperform the zero-shot base VLM, although they remain below the full \memel{} model. Figure~\ref{fig:loto_lolo_transfer}(b) shows that these gains vary across tasks and languages. Generalization gains are larger when zero-shot performance is lower, and this inverse relationship holds within each setting (Spearman $\rho=-0.45$, $p=0.005$ for LOTO; $\rho=-0.62$, $p<10^{-4}$ for LOLO), suggesting that unified training is most beneficial when the base model has weaker zero-shot capability. For example, Macro-F1 improves substantially for metaphor and the fine-grained MIMIC misogyny categories. Together with the single-dataset results, these findings support generalization across tasks and languages, while task-specific label definitions may still benefit from in-domain supervision. Detailed results are provided in Appendix~\ref{appendix:loto_lolo}.

\section{Conclusion and Future Work} 
\label{sec:conclusions} 
We presented \textsc{MemeLens}, a unified multilingual, multitask, explanation-enhanced vision--language model for meme understanding. We consolidated 38 public meme datasets and harmonized their annotations into a shared taxonomy of 20 tasks spanning harm, targets, figurative and pragmatic intent, and affect. Our analyses across modeling paradigms, task categories, and datasets show that robust meme understanding benefits from multimodal training, with performance varying across semantic categories. Dataset-level analysis further reveals a trade-off between unification and specialization: unified training provides broad coverage and robustness while remaining broadly comparable to dataset-specific state-of-the-art models under our indicative evaluation.

In future work, we aim to further improve robustness in low-resource and diverse settings while expanding to more languages, cultures, and emerging meme trends, incorporating continual learning, domain adaptation, and culturally grounded evaluation protocols to better capture evolving online communication dynamics.

\section*{Limitations}

\textsc{MemeLens} consolidates public resources and therefore inherits dataset-specific artifacts, including platform biases, annotation conventions, and temporal drift (e.g., evolving slang, symbols, and community norms). As a result, performance under unified labels may not fully reflect robustness to emerging meme formats or shifting online contexts. Label unification enables cross-dataset learning and comparison, but may compress fine-grained distinctions and introduce ambiguity. Explanation supervision is generated and/or curated under practical constraints and may therefore miss relevant cues, over-emphasize salient regions, or provide incomplete rationales. Finally, although \textsc{MemeLens} covers multiple languages, coverage remains uneven across languages and tasks, particularly in low-resource settings.

\section*{Ethics and Broader Impact}
\textsc{MemeLens} uses public datasets to support multilingual and robust multimodal research. Since memes may contain hateful, offensive, or sensitive content, trained models may reproduce harmful language or stereotypes. We therefore recommend responsible use, with clear documentation, content filtering, and human oversight in high-stakes settings. Generated explanations are intended to describe evidence rather than endorse harmful messages, but may still contain sensitive concepts. While \textsc{MemeLens} can support moderation, cross-cultural analysis, and multilingual accessibility, it could also be misused for harmful purposes, motivating careful deployment and monitoring.

\section*{Acknowledgments}
The work of A. E. Shahroor, M. B. Kmainasi, G. D. S. Martino and F. Alam was supported by NPRP grant 14C-0916-210015 from the Qatar National Research Fund, part of the Qatar Research Development and Innovation Council (QRDI). The findings reported herein are solely the responsibility of the authors. The work of Dimitar Dimitrov is partially supported by the project UNITe BG16RFPR002-1.014-0004 funded by PRIDST.

\bibliography{bibliography/main}

\appendix

\newpage

\begin{table*}[!tbh]
\centering
\resizebox{0.94\textwidth}{!}{%
\setlength{\tabcolsep}{4pt}
\begin{tabular}{llcll}
\toprule
\textbf{Dataset} & \textbf{Task} & \textbf{\#} & \textbf{Original Labels} & \textbf{Canonical Labels} \\
\midrule
\multirow{3}{*}{BanglaAbuseMeme}
  & sarcasm & 2 & Yes / No & \texttt{sarcasm} / \texttt{not-sarcasm} \\
  & vulgar  & 2 & Vulgar / Not Vulgar & \texttt{vulgar} / \texttt{not-vulgar} \\
  & abuse   & 2 & Abusive / Non-abusive & \texttt{abusive} / \texttt{not-abusive} \\
\midrule
Toxic-memes (RU) & toxic & 2 & 1 / 0 & \texttt{toxic} / \texttt{non-toxic} \\
\midrule
\multirow{2}{*}{RoMemes}
  & deepfake  & 3 & Real / Fake / DeepFake & \texttt{real} / \texttt{fake} / \texttt{deepfake} \\
  & political & 2 & Yes / No & \texttt{political} / \texttt{not-political} \\
\midrule
\multirow{2}{*}{MIMIC2024}
  & misogyny & 2 & 1 / 0 & \texttt{misogynous} / \texttt{not-misogynous} \\
  & misogyny cat.$^\dagger$ & 7 & multi-label codes & \makecell[l]{\texttt{obj.} / \texttt{prej.} / \texttt{hum.} / \texttt{unspec.} \\ \texttt{obj.,prej.} / \texttt{obj.,hum.} / \texttt{prej.,hum.}} \\
\midrule
Multi3Hate (en/de/zh/es/hi) & hateful & 2 & 1 / 0 & \texttt{hateful} / \texttt{not-hateful} \\
\midrule
MMHS$^\dagger$ & hateful & 2 & any label containing ``hate'' / otherwise & \texttt{hateful} / \texttt{not-hateful} \\
\midrule
\multirow{4}{*}{Memotion}
  & humour       & 4 & not\_funny / funny / very\_funny / hilarious & \texttt{not-funny} / \texttt{funny} / \texttt{very-funny} / \texttt{hilarious} \\
  & sarcasm      & 4 & \makecell[l]{not\_sarcastic / general / twisted\_meaning /\\ very\_twisted} & \makecell[l]{\texttt{not-sarcastic} / \texttt{general-sarcasm} / \\\texttt{twisted-meaning} / \texttt{very-twisted}} \\
  & motivational & 2 & motivational / not\_motivational & \texttt{motivational} / \texttt{not-motivational} \\
  & offensive    & 4 & \makecell[l]{not\_offensive / slight / very\_offensive / \\hateful\_offensive} & \makecell[l]{\texttt{not-offensive} / \texttt{sl.-offensive} / \\\texttt{very-offensive} / \texttt{hateful-offensive}} \\
\midrule
\multirow{3}{*}{MET-Meme (En)}
  & intention$^\dagger$ & 4 & 1 / 2 / 3 / 4 / 5 $\to$ removed \textit{other} & \makecell[l]{\texttt{expressive} / \texttt{offensive} / \\\texttt{interactive} / \texttt{entertaining}} \\
  & offensive   & 4 & 0 / 1 / 2 / 3 & \makecell[l]{\texttt{not-offensive} / \texttt{sl.-offensive} / \\ \texttt{mod.-offensive} / \texttt{very-offensive}} \\
  & metaphor    & 2 & 1 / 0 & \texttt{metaphorical} / \texttt{literal} \\
\midrule
\multirow{3}{*}{MET-Meme (Zh)}
  & intention   & 5 & 1 / 2 / 3 / 4 / 5 & \makecell[l]{\texttt{expressive} / \texttt{offensive} / \texttt{interactive} / \\\texttt{entertaining} / \texttt{other}} \\
  & offensive   & 4 & 0 / 1 / 2 / 3 & \makecell[l]{\texttt{not-offensive} / \texttt{sl.-offensive} / \\\texttt{mod.-offensive} / \texttt{very-offensive}} \\
  & metaphor    & 2 & 1 / 0 & \texttt{metaphorical} / \texttt{literal} \\
\midrule
\multirow{2}{*}{HarMeme$^\dagger$}
  & harmful & 3 & \makecell[l]{not\_harmful / partially\_harmful / \\very\_harmful (removed \textit{unspec.})} & \makecell[l]{\texttt{not-harmful} / \texttt{partially-harmful} / \\ \texttt{very-harmful}} \\
  & target  & 5 & none / individual / org. / comm. / society & \texttt{none} / \texttt{indiv.} / \texttt{org.} / \texttt{comm.} / \texttt{society} \\
\midrule
\multirow{2}{*}{HarMeme-Covid}
  & harmful & 3 & not\_harmful / partially\_harmful / very\_harmful & \makecell[l]{\texttt{not-harmful} / \texttt{partially-harmful} / \\\texttt{very-harmful}} \\
  & target  & 5 & none / individual / org. / comm. / society & \texttt{none} / \texttt{indiv.} / \texttt{org.} / \texttt{comm.} / \texttt{society} \\
\midrule
\multirow{5}{*}{MAMI}
  & misogynous      & 2 & 1 / 0 & \texttt{misogynous} / \texttt{not-misogynous} \\
  & shaming         & 2 & 1 / 0 & \texttt{shaming} / \texttt{not-shaming} \\
  & stereotype      & 2 & 1 / 0 & \texttt{stereotype} / \texttt{not-stereotype} \\
  & obj.            & 2 & 1 / 0 & \texttt{obj.} / \texttt{not-obj.} \\
  & violence        & 2 & 1 / 0 & \texttt{violence} / \texttt{not-violence} \\
\midrule
ArMeme$^\dagger$ & propaganda & 2 & propaganda / not\_propaganda (4 $\to$ 2) & \texttt{propaganda} / \texttt{not-propaganda} \\
\midrule
FHM & hateful & 2 & {hateful / not-hateful} & \texttt{hateful} / \texttt{not-hateful} \\
\midrule
Prop2Hate-Meme & hateful & 2 & {hateful / not-hateful} & \texttt{hateful} / \texttt{not-hateful} \\
\midrule
MIMIC-Islamophobia & hateful & 2 & 1 / 0 & \texttt{hateful} / \texttt{not-hateful} \\
\midrule
MUTE & hateful & 2 & hate / not-hate & \texttt{hateful} / \texttt{not-hateful} \\
\bottomrule
\end{tabular}%
}
\vspace{-0.3cm}
\caption{Label mapping from original annotations to canonical labels in \textsc{MemeLens}. \textbf{\#}\,=\,number of canonical classes. $\dagger$\,=\,original label set was reduced. Abbreviations: \emph{obj.}\,=\,objectification, \emph{prej.}\,=\,prejudice, \emph{hum.}\,=\,humiliation, \emph{unspec.}\,=\,unspecified, \emph{sl.}\,=\,slightly, \emph{mod.}\,=\,moderately, \emph{org.}\,=\,organization, \emph{comm.}\,=\,community, \emph{indiv.}\,=\,individual.}
\label{tab:label-mapping}
\vspace{-0.35cm}
\end{table*}
\section{Label Taxonomy}
\label{sec_app_label_taxonomy}
Table~\ref{tab:label-mapping} summarises the full label harmonisation used to construct \memel{}, mapping each dataset-specific annotation scheme to the canonical labels adopted in our unified taxonomy. The constituent datasets adopt heterogeneous label schemes, ranging from simple binary indicators such as \texttt{0/1} or \texttt{Yes/No}, to fine-grained multi-class severity rubrics (e.g., MET-Meme offensiveness levels), structured target categories (e.g., HarMeme), and multi-label codes for misogyny subtypes (e.g., MIMIC2024). To enable consistent cross-dataset training and evaluation, we map each original label to a human-readable canonical form following a uniform \texttt{positive}/\texttt{not-positive} naming convention (e.g., \texttt{hateful}/\texttt{not-hateful}, \texttt{vulgar}/\texttt{not-vulgar}). 

For datasets whose original label spaces are reduced, marked with $\dagger$ in the table, we either map rare or ambiguous subclasses into the canonical task label (e.g., the 41-way hate subtypes in MMHS are mapped to \texttt{hateful}/\texttt{not-hateful}) or remove under-represented categories (e.g., the \texttt{other} class in MET-Meme English and the \texttt{unspecified} class in HarMeme). Importantly, the taxonomy is \emph{not} uniformly binary, since we preserve the full cardinality of the original label space when it is semantically meaningful and stable across benchmarks. Examples include the 3-way deepfake labels in RoMemes, the 3-way harmfulness and 5-way target schemes in HarMeme and HarMeme-COVID, the 4-way humour, sarcasm, and offensiveness scales in Memotion, and the 7-way multi-label misogyny categories in MIMIC2024, which we represent by decomposing composite codes (e.g., \texttt{prejudice+humiliation}) into shared primitives. Overall, this mapping policy minimises spurious variation introduced by annotation conventions, while preserving genuinely distinct semantic distinctions across datasets, tasks, and languages.

\section{Explanation Quality}
We assess the quality of the generated explanations using four established criteria, namely \emph{informativeness}, \emph{clarity}, \emph{plausibility}, and \emph{faithfulness}. Each criterion is scored on a 1--5 Likert scale, following the evaluation setup reported in \citet{kmainasi-etal-2025-memeintel}.
Since human annotation at the full \memel{} scale (38 datasets, $\sim$40K test instances) is prohibitively expensive, we complement targeted human review with a large-scale automatic evaluation using two strong and independent LLM judges, \textit{Gemini 2.5 Pro}~\cite{comanici2025gemini} and \textit{GPT-5}~\cite{singh2025openai}, in order to reduce single-judge bias. Table~\ref{tab:judge_scores} reports the LLM-judge results aggregated over the entire test set. Both judges assign consistently high scores across all four criteria, indicating that the generated explanations are generally coherent, informative, and well grounded in the multimodal input. GPT-5 yields a higher overall mean than Gemini 2.5 Pro (4.65 vs.\ 4.36), with the higher gaps appearing in \emph{clarity} (4.80 vs.\ 4.57) and \emph{faithfulness} (4.69 vs.\ 4.30). However, the ranking of criteria is consistent across the two judges, providing evidence that the explanations capture the relevant cues underlying each label.

\begin{table}[!tbh]
\centering
\setlength{\tabcolsep}{6pt}
\scalebox{0.8}{
\begin{tabular}{lccccc}
\toprule
\textbf{Model} & \textbf{Info.} & \textbf{Clar.} & \textbf{Plaus.} & \textbf{Faith.} & \textbf{Overall} \\
\midrule
Gemini-2.5 & 4.35 & 4.57 & 4.24 & 4.30 & 4.36 \\
GPT-5      & 4.48 & 4.80 & 4.62 & 4.69 & 4.65 \\
\bottomrule
\end{tabular}
}
\caption{LLM-judge scores for explanation quality on the whole test set ($\sim$40K). \textbf{Info.} = informativeness, \textbf{Clar.} = clarity, \textbf{Plaus.} = plausibility, and \textbf{Faith.} = faithfulness. \textbf{Gemini-2.5} = Gemini 2.5 Pro.}
\label{tab:judge_scores}
\vspace{-0.3cm}
\end{table}

To validate the automatic evaluation, Table~\ref{tab:human_metrics} reports a human evaluation on a stratified subset drawn from four representative datasets spanning different languages and tasks: 
ArMeme (Arabic propaganda), FHM (English hate), Multi3Hate (multilingual hate), and MIMIC (Hindi--English misogyny). Although the absolute human scores are systematically lower than those of the LLM judges, the relative trend is broadly consistent. FHM receives the strongest and most balanced ratings across all criteria, while ArMeme is clearly the most difficult, particularly for \emph{plausibility} (2.89) and \emph{faithfulness} (2.85). Multi3Hate and MIMIC show moderate, comparable performance. Across datasets, \emph{clarity} is typically rated higher than \emph{plausibility} and \emph{faithfulness}, suggesting that the model produces explanations that are easy to follow, with further room to strengthen their grounding in the multimodal evidence. Overall, the LLM-judge and human evaluations indicate that the explanation supervision in \memel{} is of high overall quality, while also highlighting that explanation quality varies substantially by dataset and remains most challenging in low-resource 
settings.

\begin{table}[!tbh]
\centering
\setlength{\tabcolsep}{3pt}
\scalebox{0.8}{
\begin{tabular}{lccccc}
\toprule
\textbf{Dataset} & \textbf{Inf.} & \textbf{Clar.} & \textbf{Plaus.} & \textbf{Faith.} & \textbf{Overall} \\
\midrule
ArMeme & 3.09 & 3.48 & 2.89 & 2.85 & 3.08 \\
FHM    & 4.67 & 4.65 & 4.64 & 4.60 & 4.64 \\
Multi3 & 3.33 & 3.95 & 3.41 & 3.33 & 3.50 \\
MIMIC  & 3.44 & 4.05 & 3.42 & 3.41 & 3.58 \\
\bottomrule
\end{tabular}
}
\caption{Human evaluation of generated explanations on a small sample from each dataset.}
\label{tab:human_metrics}

\end{table}

\section{Dataset Details}
\label{sec_app_dataset}

Table~\ref{tab:dataset_stats} summarizes the data distribution across tasks, datasets, and languages. The benchmark covers binary, multi-class, and multi-label classification, with 178K training, 22K validation, and 40K test instances overall. We briefly describe each dataset below.

\begin{table*}[!tbh]
\centering
\small
\sisetup{
  input-ignore = {,},
  group-separator = {,},
  group-minimum-digits = 4
}
\begin{tabular}{@{}llll S[table-format=6] S[table-format=5] S[table-format=5]@{}}
\toprule
\textbf{Task} & \textbf{Dataset} & \textbf{Language} & \textbf{Label Type} & {\textbf{Train}} & {\textbf{Val}} & {\textbf{Test}} \\ 
\midrule
Abuse & BanglaAbuseMeme & Bengali & Binary & 2,827 & 404 & 806 \\
Deepfake & RoMemes & Romanian & Multi-class & 322 & 47 & 93 \\
Harmful & HarMeme (COVID-19) & English & Multi-class & 3,008 & 174 & 354 \\
Harmful & HarMeme & English & Multi-class & 2,937 & 176 & 355 \\
Hateful & FHM & English & Binary & 8,500 & 540 & 2,000 \\
Hateful & MIMIC (Islamophobia) & English & Binary & 515 & 75 & 150 \\
Hateful & MMHS & English & Binary & 41,484 & 5,887 & 11,881 \\
Hateful & Prop2Hate-Meme & Arabic & Binary & 2,143 & 312 & 606 \\
Hateful & MUTE & Bengali & Binary & 3,365 & 375 & 416 \\
Hateful & Multi3Hate & German & Binary & 209 & 30 & 61 \\
Hateful & Multi3Hate & English & Binary & 209 & 30 & 61 \\
Hateful & Multi3Hate & Spanish & Binary & 209 & 30 & 61 \\
Hateful & Multi3Hate & Hindi & Binary & 209 & 30 & 61 \\
Hateful & Multi3Hate & Chinese & Binary & 209 & 30 & 61 \\
Humor & Memotion & English & Multi-class & 4,890 & 699 & 1,398 \\
Intention & MET-Meme & English & Multi-class & 2,749 & 396 & 784 \\
Intention & MET-Meme & Chinese & Multi-class & 4,067 & 584 & 1,161 \\
Metaphor & MET-Meme & English & Binary & 2,754 & 395 & 788 \\
Metaphor & MET-Meme & Chinese & Binary & 4,061 & 587 & 1,166 \\
Misogyny & MAMI & English & Binary & 9,000 & 1,000 & 1,000 \\
Misogyny & MIMIC2024 & Hindi-English & Binary & 3,448 & 490 & 967 \\
Misogyny (Cat.) & MIMIC2024 & Hindi-English & Multi-label & 3,429 & 486 & 988 \\
Motivational & Memotion & English & Binary & 4,890 & 700 & 1,397 \\
Objectification & MAMI & English & Binary & 9,000 & 1,000 & 1,000 \\
Offensive & Memotion & English & Multi-class & 4,888 & 700 & 1,399 \\
Offensive & MET-Meme & English & Multi-class & 2,752 & 396 & 789 \\
Offensive & MET-Meme & Chinese & Multi-class & 4,064 & 580 & 1,170 \\
Political & RoMemes & Romanian & Binary & 322 & 47 & 93 \\
Propaganda & ArMeme & Arabic & Binary & 3,604 & 522 & 1,021 \\
Sarcasm & Memotion & English & Multi-class & 4,888 & 700 & 1,399 \\
Sarcasm & BanglaAbuseMeme & Bengali & Binary & 2,827 & 404 & 806 \\
Shaming & MAMI & English & Binary & 9,000 & 1,000 & 1,000 \\
Stereotype & MAMI & English & Binary & 9,000 & 1,000 & 1,000 \\
Target (COVID) & HarMeme & English & Multi-class & 3,008 & 174 & 354 \\
Target & HarMeme & English & Multi-class & 2,938 & 176 & 355 \\
Toxic & Toxic Memes & Russian & Binary & 4,512 & 647 & 1,297 \\
Violence & MAMI & English & Binary & 9,000 & 1,000 & 1,000 \\
Vulgar & BanglaAbuseMeme & Bengali & Binary & 2,825 & 405 & 807 \\
\midrule
\textbf{Total} & & & & \textbf{178,062} & \textbf{22,228} & \textbf{40,105} \\
\bottomrule
\end{tabular}
\vspace{-0.2cm}
\caption{Data distribution across tasks, datasets, and languages.}
\label{tab:dataset_stats}
\vspace{-0.3cm}
\end{table*}

\paragraph{BanglaAbuseMeme~\cite{das-etal-2023-banglaabusememe}.}
This is a Bengali (Bangla) multimodal meme dataset curated from social platforms, where each example pairs an image with its embedded text (using OCR) for abusive-language understanding. It is designed to support abusive content detection in memes, emphasizing cases where abusiveness emerges from the interaction of image and text. We use the \emph{abuse}, \emph{sarcasm}, and \emph{vulgarity} subsets of BanglaAbuseMeme, formulating each as a \textbf{binary classification task} (present vs.\ not-present).

\paragraph{RoMemes~\cite{puaics2024romemes}.}
RoMemes is a Romanian-language multimodal meme benchmark consisting of image--text pairs annotated for multiple meme understanding tasks. The dataset supports evaluation beyond high-resource English settings by focusing on Romanian social media content. In our experiments, we use the \emph{political meme detection} subset and the \emph{fake image detection}.

\paragraph{HarMeme~\cite{pramanick-etal-2021-detecting}.}
HarMeme is an English-language multimodal meme dataset annotated for harmfulness and target identification, where each meme image is paired with embedded textual content. The dataset provides fine-grained harmfulness labels (e.g., not harmful, partially harmful, very harmful) and annotations indicating the targeted entity (e.g., individual, organization, group). HarMeme is organized into two subsets: Harm-C, consisting of COVID-19-related memes, and Harm-P, consisting of political memes from the U.S.\ context. In our experiments, we use Harm-C from the original HarMeme-V0 release and Harm-P from the updated HarMeme-V1 release. In addition to harmfulness, we use the target annotation; memes without an explicit target are treated as \emph{none}, following the dataset schema.

\paragraph{Islamophobic memes dataset~\cite{islam2024mimic}.}
The Islamophobic memes dataset is a curated collection of multimodal memes designed to study Islamophobic and anti-Muslim content in online discourse. Each instance consists of a meme image paired with its embedded textual content, with annotations supporting supervised detection of Islamophobic (hateful) versus non-hateful memes. In our experiments, we exclude samples whose images do not contain embedded text, focusing on instances where both visual and textual modalities contribute to the expression of hateful intent.

\paragraph{MMHS150K~\cite{Gomez2020exploring}.}
MMHS150K is a large-scale multimodal hate-speech dataset of social media posts that pairs images with accompanying textual content. The dataset is designed with the aim to support hate-speech detection in multimodal settings, including cases where the hateful meaning emerges from the interaction between the visual and the textual signals. In our experiments, we exclude any examples for which the images contain no embedded text, and we focus on instances where both modalities actually contribute to the multimodal context.

\paragraph{Multi3Hate~\cite{bui-etal-2025-multi3hate}.}
Multi3Hate is a multilingual, parallel meme dataset created by instantiating the same meme templates across multiple languages, enabling controlled cross-lingual evaluation of hate-speech detection. By keeping visual content fixed while varying language realizations, it supports analysis of cross-lingual robustness and transfer in multimodal hate detection. In our experiments, we use the German, English, Spanish, Hindi, and Chinese subsets of the dataset.

\paragraph{Prop2Hate~\cite{alam2024propaganda}.}
Prop2Hate is an Arabic multimodal meme dataset constructed with the aim to study the intersection between propagandistic and hateful content in memes. The dataset extends an existing Arabic propagandistic meme collection by annotating memes for hatefulness, where each instance pairs a meme image with its embedded textual content. The annotations in this dataset support supervised hate-speech detection in multimodal settings, capturing cases where hateful meaning emerges from the interaction between visual and textual cues. In our experiments, we use the dataset in a \textbf{binary classification} setup, distinguishing between \emph{hateful} and \emph{non-hateful} memes.

\paragraph{Memotion~\cite{sharma-etal-2020-semeval}.}
Memotion is a multimodal meme dataset released as part of SemEval-2020 Task 8. The dataset pairs meme images with their corresponding embedded textual content. The dataset provides annotations for multiple emotion-related categories, including \emph{humor}, \emph{sarcasm}, \emph{offensiveness}, and \emph{motivation}, thus enabling the study of affective phenomena in memes that arise from the interaction between visual and the textual cues.

\paragraph{MUTE~\cite{hossain-etal-2022-mute}.}
MUTE is a Bengali multimodal meme dataset introduced to support hateful meme detection in low-resource language settings. The dataset pairs meme images with their textual content, including both Bengali and Bengali--English code-mixed captions, and provides annotations for supervised hate-speech detection. In our experiments, we use the Bengali subset of MUTE under a \textbf{binary classification} setup with two labels (\emph{hateful} vs.\ \emph{non-hateful}).

\paragraph{MET-Meme~\cite{xu-2022-metmeme}.}
MET-Meme is a metaphor-rich multimodal meme dataset that pairs images with their textual content to support multimodal meme understanding. The dataset provides annotations for multiple semantic aspects of memes, including metaphor occurrence, intention, and offensiveness, capturing cases where meaning arises from the interaction between visual and textual elements. In our experiments, we use the \emph{metaphor occurrence detection}, \emph{intention detection}, and \emph{offensiveness detection} annotations for both the English and Chinese subsets of the dataset.

\paragraph{MAMI~\cite{fersini-etal-2022-semeval}.}
MAMI is a dataset released for SemEval-2022 Task 5, consisting of meme images paired with embedded textual content. The dataset provides annotations for misogyny at both a coarse level (misogynous vs.\ non-misogynous) and a fine-grained level, including \emph{shaming}, \emph{stereotype}, \emph{objectification}, and \emph{violence}. In our experiments, we use misogyny identification as the primary task. Fine-grained categories (\textit{violence}, \textit{objectification}, \textit{shaming}, \textit{stereotype}) are modeled as auxiliary binary prediction tasks within our unified framework, but are not treated as independent benchmarks.

\paragraph{MIMIC~\cite{singh2024mimic}.}
MIMIC is a Hindi--English code-mixed multimodal dataset for misogyny identification in online memes and posts. Each example combines an image with short, often code-mixed textual content, and is annotated for misogyny. The dataset is provided in two variants: one formulated as a \emph{binary misogyny detection} dataset (misogynous vs.\ non-misogynous), and another containing \emph{category-level annotations} for misogyny-related subtypes. We use the category-level dataset as well, where memes are labeled with the following categories: \emph{unspecified}, \emph{prejudice}, \emph{objectification}, \emph{humiliation}, as well as their observed combinations (e.g., \emph{objectification+humiliation}, \emph{prejudice+humiliation}, \emph{objectification+prejudice}).

\paragraph{ArMeme~\cite{alam-etal-2024-armeme}.}
ArMeme is an Arabic multimodal meme dataset designed for propaganda detection. It consists of meme images paired with their overlaid text and corresponding propaganda annotations. In this work, we use the binary version of ArMeme, which includes two labels: \emph{propaganda} and \emph{not propaganda}.

\paragraph{FHM~\cite{kiela2020hatefulmemes}.}
FHM (Facebook Hateful Memes) is an English-language multimodal meme dataset introduced as part of the Hateful Memes Challenge. Each example pairs a meme image with its textual content and is annotated for hate speech under a binary labeling scheme (hateful vs.\ non-hateful). The dataset is specifically designed to require joint reasoning over visual and textual modalities by including challenging confounders that prevent reliance on unimodal cues, making it a standard benchmark for evaluating multimodal hate-speech detection models.

\paragraph{Toxic Memes Detection Dataset~\cite{raikovskaia2023toxicmemes}.}
The Toxic Memes Detection Dataset is a Russian-language multimodal meme dataset released on Zenodo, consisting of images collected from popular Russian Telegram channels and annotated for toxic content according to Facebook Community Standards. The dataset supports supervised toxic content detection in memes through a binary labeling scheme. As the dataset provides image-level annotations without explicit textual transcriptions, we extract the embedded text from meme images using OCR to construct image--text pairs for multimodal modeling.

\begin{table}[t]
\centering
\footnotesize
\setlength{\tabcolsep}{2pt}
\begin{tabular}{@{}lllr@{}}
\toprule
\textbf{Dataset} & \textbf{Task} & \textbf{Ref.} & \textbf{Score} \\
\midrule
\multirow{3}{*}{BanglaAbuse.} 
  & Vulg. (Ma-F1) & \multirow{3}{*}{\citeyear{das-etal-2023-banglaabusememe}} & 71.66 \\
  & Sarc. (Ma-F1) & & 68.28 \\
  & Abuse (Ma-F1) & & 70.51 \\
\midrule
\multirow{2}{*}{RoMemes} 
  & Deepfake (Acc.) & \multirow{2}{*}{\citeyear{puaics2024romemes}} & 79.7 \\
  & Political (Acc.) & & 62.0 \\
\midrule
\multirow{2}{*}{HarMeme} 
  & Harm-C (Ma-F1) & \multirow{2}{*}{\citeyear{pramanick-etal-2021-detecting}} & 53.85 \\
  & Harm-P (Ma-F1) & & 64.70 \\
\midrule
\multirow{5}{*}{Multi3Hate} 
  & ENG (Acc.) & \multirow{5}{*}{\citeyear{bui-etal-2025-multi3hate}} & 75.8 \\
  & DE (Acc.) & & 72.2 \\
  & ES (Acc.) & & 69.2 \\
  & HI (Acc.) & & 63.1 \\
  & ZH (Acc.) & & 68.7 \\
\midrule
Prop2Hate & Hate (Ma-F1) & \citeyear{alam2024propaganda} & 70.9 \\
\midrule
MAMI & Misog. Classif. (F1) & \citeyear{zhang2022srcb} & 83.4 \\
\midrule
MUTE & Classif. (W-F1) & \citeyear{hossain-etal-2022-mute} & 67.2 \\
\midrule
MIMIC & Islam. (Ma-F1) & \citeyear{islam2024mimic} & 69.5 \\
\midrule
MMHS150K & Hate (Acc.) & \citeyear{Gomez2020exploring} & 68.4 \\
\midrule
FHM & Hate (Acc.) & \citeyear{kmainasi-etal-2025-memeintel} & 79.2 \\
\midrule
\multirow{2}{*}{MET-Meme} 
  & EN (F1-pos) & \multirow{2}{*}{\citeyear{xu-2022-metmeme}} & 82.39 \\
  & ZH (F1-pos) & & 77.23 \\
\midrule
Russ. Toxic & -- & N/A & -- \\
\bottomrule
\end{tabular}
\vspace{-0.2cm}
\caption{SOTA reference results. Ma-F1: Macro-F1; W-F1: Weighted F1; Acc.: Accuracy.}
\label{tab:sota}
\vspace{-0.4cm}
\end{table}

\section{State-of-the-Art (SOTA) Results}
\label{sec:sota}

To contextualize our results, we report published state-of-the-art (SOTA) or best-reported performance for each dataset and task, as summarized in Table~\ref{tab:sota}. All numbers are taken directly from the original dataset papers or subsequent benchmark studies and are provided for reference only. We did not reproduce the published results.

\begin{table}[t]
\centering
\small
\setlength{\tabcolsep}{3pt}
\scalebox{0.9}{
\begin{tabular}{@{}llcrr@{}}
\toprule
\textbf{Dataset} & \textbf{Task} & \textbf{L} & \textbf{EN Avg} & \textbf{Nat Avg} \\
\midrule
BanglaAbuse & Abuse & BN & 116 & 93 \\
RoMemes & Deepfake & RO & 122 & 110 \\
HarMeme (Co.) & Harmful & EN & 122 & - \\
HarMeme & Harmful & EN & 125 & -\\
FHM & Hateful & EN & 110 & - \\
MIMIC\_Isl & Hateful & EN & 114 & - \\
MMHS & Hateful & EN & 122 & 112 \\
Prop2Hate & Hateful & AR & 123 & 94 \\
MUTE & Hateful & BN & 121 & 95 \\
Multi3Hate & Hateful & DE & 113 & 102 \\
Multi3Hate & Hateful & EN & 112 & - \\
Multi3Hate & Hateful & ES & 113 & 114 \\
Multi3Hate & Hateful & HI & 115 & 118 \\
Multi3Hate & Hateful & ZH & 112 & - \\
Memotion & Humor & EN & 128 & - \\
MET-Meme & Intention & EN & 122 & - \\
MET-Meme & Intention & ZH & 114 & - \\
MET-Meme & Metaphor & EN & 120 & - \\
MET-Meme & Metaphor & ZH & 113 & - \\
MAMI & Misogyny & EN & 116 & - \\
MIMIC2024 & Misogyny & HI-EN & 119 & 127 \\
MIMIC2024 (Cat.) & Misogyny & HI-EN & 123 & 128 \\
Memotion & Motiv. & EN & 114 & - \\
MAMI & Objectif. & EN & 122 & - \\
Memotion & Offensive & EN & 124 & - \\
MET-Meme & Offensive & EN & 119 & - \\
MET-Meme & Offensive & ZH & 109 & - \\
RoMemes & Political & RO & 117 & 104 \\
ArMeme & Propaganda & AR & 125 & 99 \\
Memotion & Sarcasm & EN & 125 & - \\
BanglaAbuse & Sarcasm & BN & 121 & 92 \\
MAMI & Shaming & EN & 118 & - \\
MAMI & Stereotype & EN & 120 & - \\
HarMeme (CO.) & Target & EN & 126 & - \\
HarMeme & Target & EN & 126 & - \\
Toxic Memes & Toxic & RU & 113 & 84 \\
MAMI & Violence & EN & 116 & - \\
BanglaAbuse & Vulgar & BN & 113 & 87 \\
\midrule
\textbf{Average} & - & - & \textbf{118} & \textbf{104} \\
\bottomrule
\end{tabular}
}
\vspace{-0.2cm}
\caption{Average explanation length (in words) per dataset, task, and language. English explanations are available for all datasets, while native-language (Nat. = Native) explanations are reported where applicable. L. = Language.}
\label{tab:explanation_length}
\vspace{-0.3cm}
\end{table}

\begin{table*}[t]
\centering
\footnotesize
\setlength{\tabcolsep}{2.5pt}
\scalebox{0.85}{%
\begin{tabular}{@{}llc|ccc|ccc|ccc|ccc@{}}
\toprule
& & & \multicolumn{3}{c|}{\textbf{Text-Only}} & \multicolumn{3}{c|}{\textbf{Image-Only}} & \multicolumn{3}{c|}{\textbf{MM-Seq}} & \multicolumn{3}{c|}{\textsc{\textbf{MemeLens}}} \\
\cmidrule(lr){4-6} \cmidrule(lr){7-9} \cmidrule(lr){10-12} \cmidrule(lr){13-15}
\textbf{Dataset} & \textbf{Task} & \textbf{Lang.} & Acc & Ma & W & Acc & Ma & W & Acc & Ma & W & Acc & Ma & W \\
\midrule
BanglaAbuse & Abuse & BN & 66.0& .564 & .615 & 68.0& .628 & .663 & 73.1& .698 & .723 &\textbf{ 78.7 }& \textbf{.759} & \textbf{.782} \\

RoMemes & Deepfake & RO & 63.4& .259 & .493 & 64.5& .399 & .630 & 57.5& .338 & .551 &\textbf{ 77.0}& .\textbf{491} &\textbf{ .753} \\

HarMeme (Co) & Harmful & EN & 71.2& .499 & .706 & 70.3& .443 & .677 & \textbf{81.1 }& \textbf{.546} & \textbf{.797} & 74.8& .523 & .740 \\

HarMeme & Harmful & EN & 49.9& .338 & .489 & 53.5& .362 & .527 & 59.0& .400 & .590 & \textbf{62.2}& \textbf{.467} &\textbf{ .617}  \\

Prop2Hate & Hateful & AR & 74.6& .427 & .637 & 74.3& .426 & .636 & \textbf{80.0}&\textbf{ .650 }&\textbf{ .760} & 77.2& .546 & .703  \\

MUTE & Hateful & BN & 64.2& .556 & .602 & 68.8& .659 & .682 & \textbf{73.0 }& \textbf{.710 }& \textbf{.730} & 71.9& .700 & .718 \\

Multi3Hate & Hateful & DE & 59.0& .371 & .438 & 55.7& .504 & .533 & 72.0& .710 & .720 & \textbf{75.4}&\textbf{ .731} & \textbf{.745} \\

MIMIC\_Isl & Hateful & EN & 64.7& .633 & .635 & 58.0& .576 & .577 & 51.0& .340 & .350 &\textbf{ 70.7}& \textbf{.707} & \textbf{.707} \\

MMHS & Hateful & EN & 63.1& .387 & .488 & 62.1& .495 & .561 &\textbf{ 63.0 }& .500 &\textbf{ .570 }& 61.4& \textbf{.516} & .568 \\

Multi3Hate & Hateful & EN & 57.4& .573 & .573 & 50.8& .507 & .507 & \textbf{77.0}& \textbf{.770 }& \textbf{.770} & 74.1& .735 & .734 \\

FHM & Hateful & EN & 63.3& .541 & .592 & 62.3& .507 & .567 & 76.0& .740 & .760 & \textbf{79.8}& \textbf{.782} &\textbf{ .798} \\

Multi3Hate & Hateful & ES & 55.7& .358 & .399 & 67.2& .661 & .668 & 62.0& .600 & .610 & \textbf{80.0}& .796 & .799 \\

Multi3Hate & Hateful & HI & 65.6& .579 & .618 & 57.4& .528 & .559 & \textbf{75.0}& .\textbf{750} & .\textbf{760} & 75.4& .724 & .744 \\

Multi3Hate & Hateful & ZH & 63.9& .390 & .499 & 57.4& .470 & .535 & 64.0& .610 & .640 & \textbf{77.0}& \textbf{.740} & \textbf{.765} \\

Memotion & Humor & EN &\textbf{ 35.3}& .204 & .276 & 32.5& .235 & .297 & 35.0& .250 & .310 & 35.2&\textbf{ .248 }& \textbf{.316} \\

MET-Meme & Intention & EN & 46.4& .353 & .444 & 38.3& .295 & .363 & 32.0& .220 & .340 & \textbf{52.4}& \textbf{.442} & \textbf{.514} \\

MET-Meme & Intention & ZH & 62.1& .443 & .611 & 44.3& .212 & .376 & 67.0& .470 & .660 & \textbf{71.0}& \textbf{.521} & \textbf{.701} \\

MET-Meme & Metaphor & EN & 81.0& .725 & .796 & 81.4& .724 & .797 & \textbf{87.0}& \textbf{.820} & \textbf{.870} & 86.7& .821 & .863 \\

MET-Meme & Metaphor & ZH & 84.7& .838 & .846 & 67.8& .636 & .662 & \textbf{90.0}& \textbf{.890} & \textbf{.890} & 86.6& .859 & .865 \\

MAMI & Misogyny & EN & 62.3& .620 & .620 & 62.8& .611 & .611 & 75.0& .740 & .740 & \textbf{84.9}& \textbf{.849} &\textbf{ .849} \\

MIMIC2024 & Misogyny (Cat.) & HI-EN & 47.0& .146 & .412 & 47.0& .146 & .412 & 66.0& .290 & .430 & \textbf{76.6}& \textbf{.592} & \textbf{.659} \\

MIMIC2024 & Misogyny & HI-EN & 67.3& .671 & .671 & 63.0& .246 & .570 & 85.0& .850 & .850 & \textbf{89.9}& \textbf{.899} & \textbf{.899}\\

Memotion & Motivational & EN & \textbf{64.7 }& .399 & .512 & 60.8& \textbf{.451} & .537 & 64.0& .450 & .540 & 63.7& .450 & \textbf{.545} \\

MAMI & Objectification & EN & 67.0& .503 & .590 & 73.2& .671 & .714 & 81.0& .780 & .810 & \textbf{83.5}& \textbf{.797} & \textbf{.826} \\

Memotion & Offensive & EN & 38.8& .140 & .217 & 37.1& \textbf{.227} & \textbf{.334} & \textbf{39.0}& .220 & .330 & 38.6& .215 & .325\\

MET-Meme & Offensive & EN & 74.8& .242 & .667 & 74.2& .233 & .658 & 74.0& \textbf{.310} & \textbf{.710} & \textbf{74.8}& .309 & .708 \\

MET-Meme & Offensive & ZH & 80.3& .485 & .787 & 74.2& .223 & .684 & 81.0& .500 & .790 & \textbf{83.0}& \textbf{.535} & \textbf{.819} \\

RoMemes & Political & RO & 67.7& .404 & .547 & 65.6& .524 & .613 & 83.0& .780 & .820 & \textbf{86.7}& \textbf{.834} &\textbf{ .858} \\

ArMeme & Propaganda & AR & 75.5& .639 & .734 & 73.5& .554 & .685 & \textbf{79.0}& \textbf{.690 }& \textbf{.770} & 78.9& .679 & .765 \\

BanglaAbuse & Sarcasm & BN & 63.9& .568 & .599 & 65.6& .636 & .651 & \textbf{68.0}& .660 & .670 & 67.4& \textbf{.661} & \textbf{.672} \\

Memotion & Sarcasm & EN & 50.2& .167 & .335 & 46.8& \textbf{.195} & \textbf{.352} & \textbf{51.0}& .170 & .340 & 50.1& .167 & .337\\

MAMI & Shaming & EN & 85.4& .461 & .787 & 83.4& .610 & .819 & 87.0& .710 & .870 & \textbf{89.8}& \textbf{.719} & \textbf{.883}\\

MAMI & Stereotype & EN & 66.1& .561 & .624 & 72.9& .336 & .707 & 74.0& .700 & .730 & \textbf{78.4}& \textbf{.739} & \textbf{.772}\\

HarMeme (Co) & Target & EN & 77.7& .449 & .788 & 72.9& .336 & .707 & \textbf{87.0}& .550 & .870 & 82.3& .420 & .840 \\

HarMeme & Target & EN & 48.5& .314 & .479 & 45.1& .204 & .404 & \textbf{59.0}& .350 & \textbf{.580 }& 56.2& \textbf{.493} & .565 \\

Toxic & Toxic & RU & 82.6& .493 & .771 & 83.9& .495 & .777 & 86.0& .700 & .850 & \textbf{86.6}& \textbf{.691} & \textbf{.853}\\

MAMI & Violence & EN & 85.3& .504 & .793 & 72.2& .618 & .702 & 91.0& .770 & .890 & \textbf{92.3}& \textbf{.809} & \textbf{.914} \\

BanglaAbuse & Vulgar & BN & 74.3& .680 & .740 & 74.3& .680 & .740 & 80.0& .750 & .800 & \textbf{82.7 }& \textbf{.772} & \textbf{.821} \\

\midrule
\multicolumn{3}{c}{\textbf{Average}} 
& 65.0& .460 & .590
& 62.7& .454 & .592
& 70.6& .578 & .679
& \textbf{74.1}& \textbf{.625} & \textbf{.720} \\

\bottomrule
\end{tabular}
}
\vspace{-0.2cm}
\caption{Performance across modalities and models on the \textit{MemeLens} benchmark. We compare text-only, image-only, multimodal sequence classification (MM-Seq), and \textit{MemeLens} (ours). Bold indicates the best score under the official metric. Acc, Ma, and W denote accuracy, macro-$F_1$, and weighted-$F_1$, respectively; pos denotes class-specific $F_1$. SOTA reports the best published result. Lang.\ denotes language.}
\label{tab:memelens_comparison}
\vspace{-0.3cm}
\end{table*}

\section{Explanation Length Statistics}
\label{app:explanation_stats}

In Table~\ref{tab:explanation_length}, we present the average explanation length across all 38 datasets. We measure explanation length in words for both English and native-language explanations where available.

\paragraph{Overall Statistics.}

Across all datasets, English explanations average 118 words, with lengths ranging from 109 words (MET-Meme Offensive in Chinese) to 128 words (Memotion Humor). This narrow range (19 words) indicates consistent levels of detail across tasks and languages. Native-language explanations are slightly shorter (104 words), with lengths ranging from 84 words (Toxic Memes in Russian) to 128 words (MIMIC category-level misogyny in Hindi–English code-mixed text).

\paragraph{Cross-Lingual Comparison.}
Native-language explanations are shorter than their English counterparts, with an average reduction of 14 words (12\%). The gap is most pronounced in Russian (29 words shorter for Toxic Memes), Bengali (21-26 words shorter across BanglaAbuse tasks), and Arabic (24-26 words shorter for ArMeme and Prop2Hate).

Conversely, Hindi and code-mixed Hindi-English explanations are comparable to or slightly longer than English versions, suggesting that explanation verbosity may be influenced by linguistic structure and expressiveness in the target language. Note that Chinese (ZH) native explanations were excluded from the native average calculation due to challenges in word-level tokenization for logographic writing systems, where character-based segmentation does not directly correspond to the word-level granularity used for other languages.

\section{Dataset-Level Results}
\label{appendix:dataset_results}
In Table~\ref{tab:memelens_comparison}, we provide a comprehensive comparison across text-only, image-only, and multimodal sequence classification baselines, alongside our proposed \textit{MemeLens} model, covering a wide range of tasks, languages, and label granularities. The results are presented using each dataset's official evaluation metric to ensure fair comparison. This detailed breakdown complements the aggregate analyses in the main paper and enables fine-grained inspection of model behavior across diverse multimodal reasoning scenarios.

\section{Dataset-Level Comparison with SOTA}
\label{appendix:dataset_sota}

As meme datasets vary substantially in task formulation, label spaces, and evaluation protocols, we group them by their official evaluation metrics, including Accuracy, Macro-$F_1$, and positive-class $F_1$ ($F_1$-POS), ensuring fair and consistent comparison across diverse experimental settings and benchmarks. This analysis assesses how a single unified multilingual, multitask model performs across heterogeneous meme benchmarks, rather than aiming to establish new dataset-specific state of the art. For each dataset, we report the performance of \textsc{\textbf{MemeLens}} under the official metric and compare it with the corresponding SOTA results from prior work.

To ensure fair comparison, we exclude datasets for which prior work does not clearly specify the $F_1$ variant used (e.g., Macro-$F_1$ vs.\ Weighted-$F_1$). Moreover, our unified training and evaluation pipeline filters out samples that do not contain textual content within the image, following a consistent text-over-image meme definition applied uniformly across all datasets. As a result, some dataset-level comparisons may not be strictly equivalent to prior results reported on unfiltered data or under differing preprocessing and experimental setups. We therefore treat these comparisons as indicative, focusing on relative trends and cross-dataset consistency rather than absolute differences.

\subsection{Accuracy-Based Evaluation}
\label{appendix:accuracy_results}

Table~\ref{tab:appendix_accuracy} reports dataset-level results for benchmarks evaluated using Accuracy ($\Delta = \text{SOTA} - \textsc{MemeLens}$). Across these benchmarks \textsc{\textbf{MemeLens}} matches or exceeds the cited dataset-specific results, with an average $\Delta$ of $-5.4$ points (i.e., in favour of \memel{}), though the margin varies widely across datasets. Because our preprocessing and evaluated instances differ from those of some published studies, these comparisons are \emph{indicative} rather than strictly like-for-like; we use them to contextualise performance rather than to claim dataset-specific state of the art.

\begin{table}[!tbh]
\centering
\small
\setlength{\tabcolsep}{3pt} %
\scalebox{0.95}{%
\begin{tabular}{lccc}
\toprule
\textbf{Dataset} & \textsc{\textbf{MemeLens}} & \textbf{SOTA} & $\boldsymbol{\Delta}$ \\
\midrule
Hateful\_de\_Multi3Hate     & 75.4 & 72.2 & -3.2 \\
Deepfake\_ro\_RoMemes      & 77.0 & 79.7 & +2.7 \\
Hateful\_en\_MMHS          & 61.4 & 68.4 & +7.0 \\
Hateful\_en\_Multi3Hate    & 74.1 & 75.8 & +1.7 \\
Hateful\_en\_FHM           & 79.8 & 79.2 & -0.6 \\
Hateful\_es\_Multi3Hate    & 80.0 & 69.2 & -10.8 \\
Hateful\_hi\_Multi3Hate    & 75.4 & 63.1 & -12.3 \\
Hateful\_zh\_Multi3Hate    & 77.0 & 68.7 & -8.3 \\
Political\_ro\_RoMemes     & 86.7 & 62.0 & -24.7 \\
\midrule
\textbf{Average} &  &  & \textbf{-5.4} \\
\bottomrule
\end{tabular}
}
\vspace{0.2cm}
\caption{Dataset-level comparison on Accuracy-based benchmarks. Accuracy is reported as a percentage (\%). $\Delta = \text{SOTA} - \textsc{\textbf{MemeLens}}$ (percentage points).}
\label{tab:appendix_accuracy}
\vspace{0.3cm}
\end{table}

\subsection{$F_1$-POS Evaluation}
\label{appendix:f1pos_results}

Table~\ref{tab:appendix_f1pos} reports results for datasets evaluated using F1 on the positive class. Across these benchmarks, \textbf{MEMELENS} achieves competitive performance relative to previously reported results, with a small average absolute difference ($\boldsymbol{\Delta} \approx 0.02$).

\begin{table}[!tbh]
\centering
\setlength{\tabcolsep}{3pt} %
\scalebox{0.85}{%
\begin{tabular}{lccc}
\toprule
\textbf{Dataset} & \textsc{\textbf{MemeLens}} & \textbf{SOTA} & $\boldsymbol{\Delta}$ \\
\midrule
Metaphor\_en\_MET & 0.729 & 0.82 & 0.09 \\
Metaphor\_zh\_MET & 0.828 & 0.77 & -0.06 \\
\midrule
\textbf{Average} &  &  & \textbf{0.02} \\
\bottomrule
\end{tabular}
}
\vspace{0.2cm}
\caption{Dataset-level comparison on $F_1$-POS benchmarks. $\Delta = \text{SOTA} - \textsc{\textbf{MemeLens}}$.}
\label{tab:appendix_f1pos}
\vspace{0.3cm}
\end{table}

\subsection{Macro-$F_1$ Evaluation}
\label{appendix:macro_results}
Table~\ref{tab:appendix_macro} reports results for datasets evaluated using Macro-$F_1$. Under this metric, \textsc{\textbf{MemeLens}} achieves performance on par with prior SOTA, suggesting that unified multitask training can match dataset-specific models even under class-imbalance-sensitive evaluation.
\begin{table}[h]
\centering
\setlength{\tabcolsep}{3pt} %
\scalebox{0.85}{%
\begin{tabular}{lccc}
\toprule
\textbf{Dataset} & \textsc{\textbf{MemeLens}} & \textbf{SOTA} & $\boldsymbol{\Delta}$ \\
\midrule
Abuse\_bn\_Bangla & 0.759 & 0.71 & -0.05 \\
Harmful\_Covid\_en & 0.523 & 0.54 & +0.02 \\
Harmful\_en\_HarMeme & 0.467 & 0.65 & +0.18 \\
Hateful\_ar\_Prop2Hate & 0.546 & 0.71 & +0.16 \\
Hateful\_en\_MIMIC & 0.707 & 0.70 & -0.01 \\
Intention\_en\_MET & 0.442 & 0.42 & -0.03 \\
Intention\_zh\_MET & 0.521 & 0.55 & +0.03 \\
Misogynous\_en\_MAMI & 0.849 & 0.83 & -0.02 \\
MisogynyCat\_hi\_MIMIC & 0.592 & 0.53 & -0.07 \\
Misogyny\_hi\_MIMIC & 0.899 & 0.73 & -0.17 \\
Sarcasm\_bn\_Bangla & 0.661 & 0.68 & +0.02 \\
Vulgar\_bn\_Bangla & 0.772 & 0.72 & -0.06 \\
Hateful\_bn\_MUTE & 0.700 & 0.67 & -0.03 \\
\bottomrule
\end{tabular}
}
\caption{Dataset-level comparison on Macro-$F_1$ benchmarks. $\Delta = \text{SOTA} - \textsc{\textbf{MemeLens}}$.}
\label{tab:appendix_macro}
\end{table}

\section{Prompts}
\label{sec_propmpts}
Tables~\ref{tab:app_meme_expl_user_prompt_en} and~\ref{tab:app_meme_expl_user_prompt_non_en} show the user-prompt templates for English-only and non-English datasets, respectively, while Table~\ref{tab:app_meme_expl_prompt} presents the system prompt used to elicit explanations for memes given an assigned label. The prompt explicitly frames the task as \emph{justification} rather than prediction. It enforces explanation quality by requiring 4--6 sentences with grounding in at least two visual cues and one textual cue, a causal link to the label (and rubric, when available), and faithful, non-speculative language. Finally, it incorporates safety and consistency constraints, including neutral handling of sensitive content and meaning-preserving bilingual explanations when a native language is requested.

\begin{table}[t]
\centering
\begin{lstlisting}[style=aclprompt]
user_prompt = f"""
TASK
- Task name: {task}
- {task_definition}

LABEL SET / RUBRIC (use this to justify the assigned label)
{labels}

MEME
- Assigned label: {label}
- Meme text (verbatim): "{text}"

OUTPUT LANGUAGE
{output_language_instruction}

Write a gold-standard explanation that justifies why this meme matches the assigned label using evidence from BOTH the image and the meme text. Return only the JSON object.
"""
\end{lstlisting}
\vspace{-0.2cm}
\caption{User prompt template for English only datasets, providing task context, meme text, and output language constraints for explanation generation.}
\label{tab:app_meme_expl_user_prompt_en}
\vspace{0.3cm}
\end{table}

\begin{table}[!tbh]
\centering
\begin{lstlisting}[style=aclprompt]
user_prompt = f"""
TASK
- Task name: {task}
- {task_definition}

LABEL SET / RUBRIC (use this to justify the assigned label)
Labels in English: {labels}
Labels in Native Language: {native_labels}

MEME
- Assigned label: {label}
- Meme text (verbatim): "{text}"

OUTPUT LANGUAGE
{output_language_instruction}

Write a gold-standard explanation that justifies why this meme matches the assigned label using evidence from BOTH the image and the meme text. Return only the JSON object.
"""
\end{lstlisting}
\vspace{-0.2cm}
\caption{User prompt template for non-English datasets, providing task context, bilingual label rubric, meme text, and output language constraints for explanation generation.}
\label{tab:app_meme_expl_user_prompt_non_en}
\vspace{-0.3cm}
\end{table}

\begin{table*}[!tbh]
\centering
\begin{lstlisting}[style=aclprompt]
system_prompt = """
You are an expert annotator for multimodal meme analysis. Your job is to write gold-standard explanations that justify a GIVEN (already assigned) label for a meme using evidence from the meme's image and its text.

You are NOT predicting the label. You must justify the provided label.

You will receive for each meme:
1) Task name
2) Task definition (what the task about) and explanation specifics (what to consider for this task)
3) The meme image
4) The meme text (verbatim, as provided)
5) The assigned label (and possibly a label definition/rubric)
6) The requested output language(s)

CRITICAL OUTPUT RULES
- Output MUST be a single valid JSON object and nothing else (no markdown, no extra keys, no commentary).
- Do NOT include line breaks inside JSON string values. Use normal spaces between sentences.
- Do not output any leading or trailing text outside the JSON object.
- If English only is requested, output exactly:
{"en_explanation":"..."}
- If English + another language is requested, output exactly:
{"en_explanation":"...","native_explanation":"..."}
- Escape any double quotes inside explanations using a backslash (\\") so the JSON remains valid.

GOLD EXPLANATION QUALITY REQUIREMENTS (apply to each requested language)
- Length: 4 to 6 sentences (prefer approximately 100 words; acceptable range 80 to 120 words).
- Evidence: explicitly reference (a) at least TWO concrete visual elements AND (b) at least ONE concrete textual element (quote a short phrase from the meme text when helpful).
- Reasoning: connect those visual + textual cues directly to why the assigned label fits (use the label definition/rubric if provided).
- Interaction: explain how image and text work together (reinforce, contrast, irony/sarcasm, punchline, etc.).
- Be precise and faithful: do not invent details that are not visible or not in the provided text. If unclear, describe generally but accurately.
- Be objective and analytical; do not endorse the meme's message.
- Sensitive content: describe neutrally. If the meme contains slurs/profanity, do not repeat them verbatim; replace them with placeholders like [SLUR] or [PROFANITY].
- If a label definition is provided, explicitly align at least one sentence with a key phrase/concept from that definition.
- Avoid generic statements like "This is funny" or "This is offensive" unless you explain exactly which visual/textual cue makes it so.

TRANSLATION CONSISTENCY (when requested)
- The native_explanation must preserve the same meaning and evidence as en_explanation (no added/removed claims).

INTERNAL SELF-CHECK (silent)
- >=2 visual cues? >=1 textual cue? 4 to 6 sentences? Clear link to label? Exact JSON keys only? Valid JSON?
"""
\end{lstlisting}
\vspace{-0.2cm}
\caption{System prompt used to elicit reference multimodal meme explanations.}
\label{tab:app_meme_expl_prompt}
\vspace{-0.3cm}
\end{table*}

\section{Experimental Details}
\label{sec_app_exp_details}

We use AdamW with cosine scheduling and a 5\% warm-up. Stage~I trains for 3 epochs with a learning rate of $1\times10^{-4}$, batch size 4 per device on 4 GPUs, and max gradient norm 1.0. We fix the seed to 42, use data-parallel training in \texttt{bfloat16}, and apply LoRA to all linear layers with $r{=}16$, $\alpha{=}32$, and dropout 0.05, keeping the vision encoder and alignment modules frozen. We evaluate after each epoch and select the best model by validation loss. Stage~II trains for 6 epochs with a learning rate of $1\times10^{-5}$.
For the \texttt{seq\_cls} baseline, we use the same setup, but train a dataset-specific head adapter for 20 epochs with learning rate $1\times10^{-5}$.

\section{Qualitative Examples}
\label{app:error-analysis}

To complement the aggregate results in Section~\ref{sec:experiments}, we present three representative cases highlighting: \textit{(i)} how explanation-augmented training (\textit{Stage~II}) corrects errors made by classification-only training (\textit{Stage~I}), and \textit{(ii)} why several \textsc{MemeLens} tasks require genuinely multimodal reasoning beyond unimodal baselines, particularly in cases involving implicit meaning, context, and cross-modal dependencies. These cases are hand-selected to illustrate representative model behaviour rather than to quantify error rates; they are qualitative illustrations that complement the aggregate results and should not be read as a systematic error taxonomy.

\begin{figure*}[!tbh]
\centering
\includegraphics[width=0.95\linewidth]{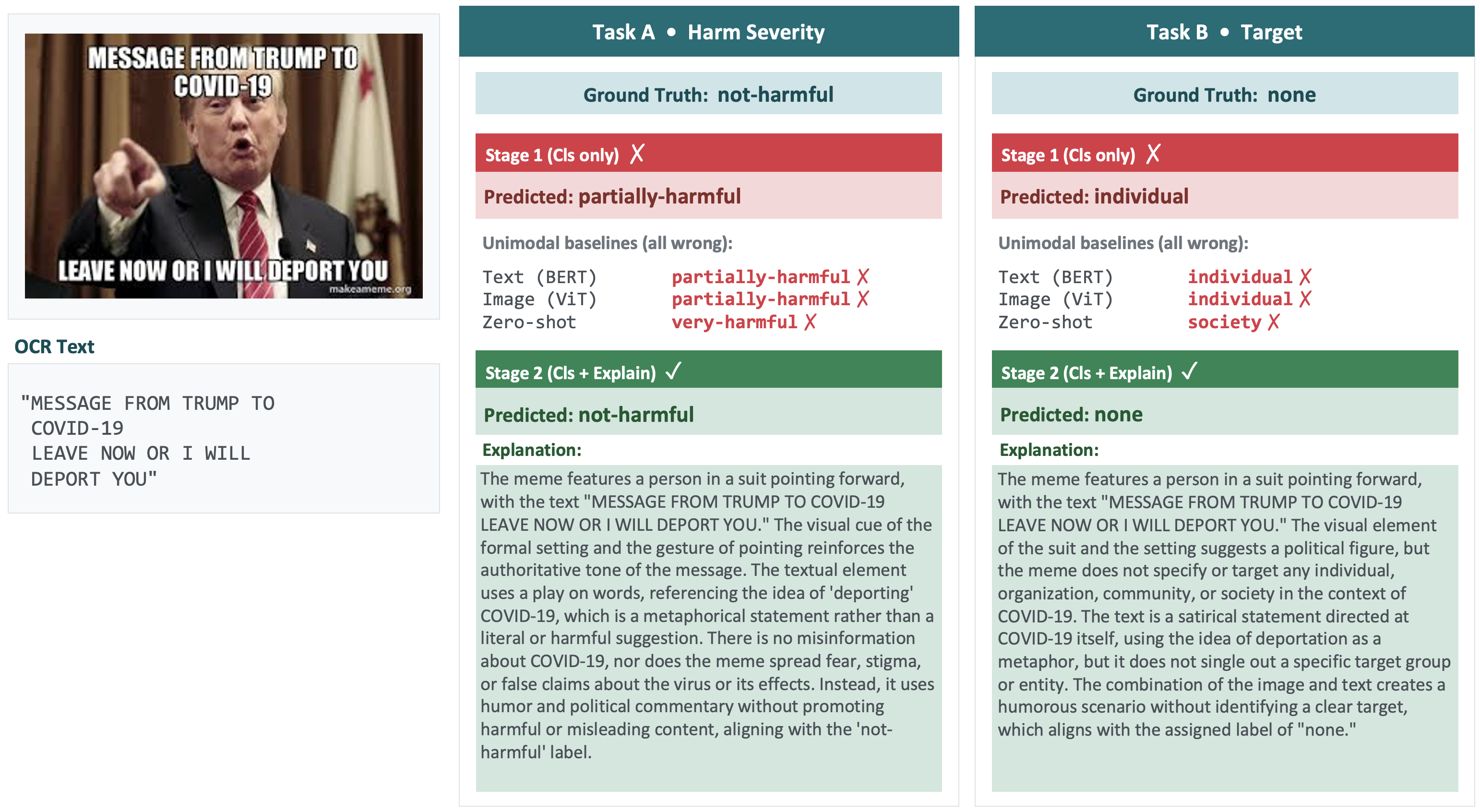}
\vspace{-0.2cm}
\caption{\textbf{Explanation corrects classification errors across schemas.} A single HarMeme instance evaluated under harm severity and target labeling. Stage~I (classification-only) fails on both tasks, while Stage~II (classification + explanation) recovers correct predictions and provides a grounded rationale.}
\label{fig:qual-expl-fix}
\vspace{-0.3cm}
\end{figure*}

\paragraph{Explanation-augmented training corrects Stage~I errors across label schemas.}
Figure~\ref{fig:qual-expl-fix} shows a COVID-related meme evaluated under two HarMeme schemas: \emph{harm severity} (Task~A) and \emph{target} (Task~B). The Stage~I model predicts \texttt{partially-harmful} and \texttt{individual}, respectively, errors also made by the text-only (BERT), image-only (ViT), and zero-shot MLLM baselines. After Stage~II training, the model correctly outputs \texttt{not-harmful} and \texttt{none}. The generated rationale attributes this correction to a metaphorical reading (``deport COVID-19'') and correctly identifies the target as the virus rather than a human entity, providing an interpretable and contextual justification grounded in multimodal evidence and task-specific annotation criteria.

\begin{figure*}[!tbh]
\centering
\includegraphics[width=0.95\linewidth]{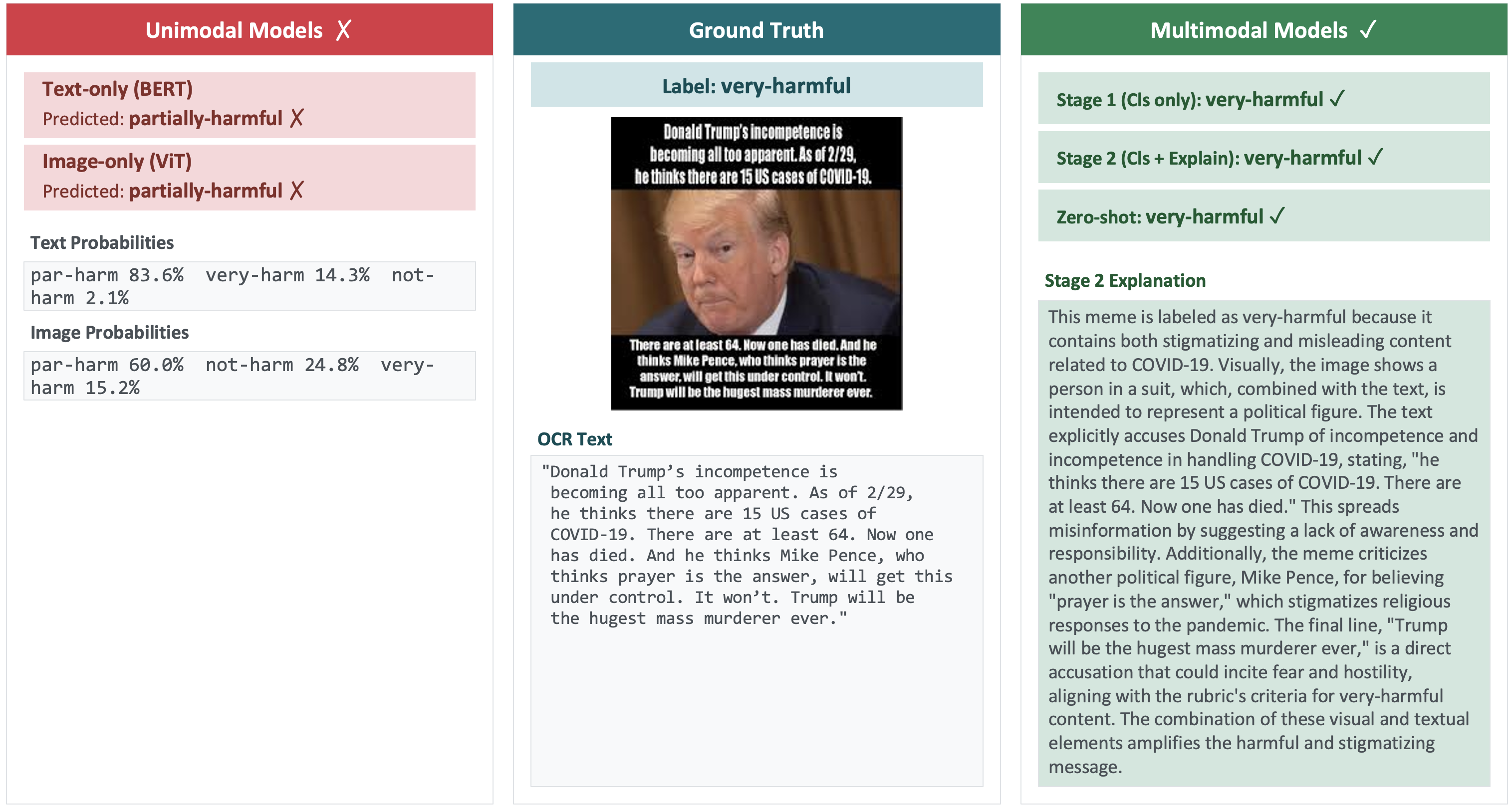}
\vspace{-0.3cm}
\caption{\textbf{Unimodal under-detection of harm.} A highly harmful meme misclassified as \texttt{partially-harmful} by unimodal baselines, but correctly identified as \texttt{very-harmful} by multimodal models. The Stage~II explanation attributes the decision to combined visual and textual cues.}
\label{fig:qual-mm-very-harmful}
\vspace{-0.3cm}
\end{figure*}

\begin{figure*}[!tbh]
\centering
\includegraphics[width=0.95\linewidth]{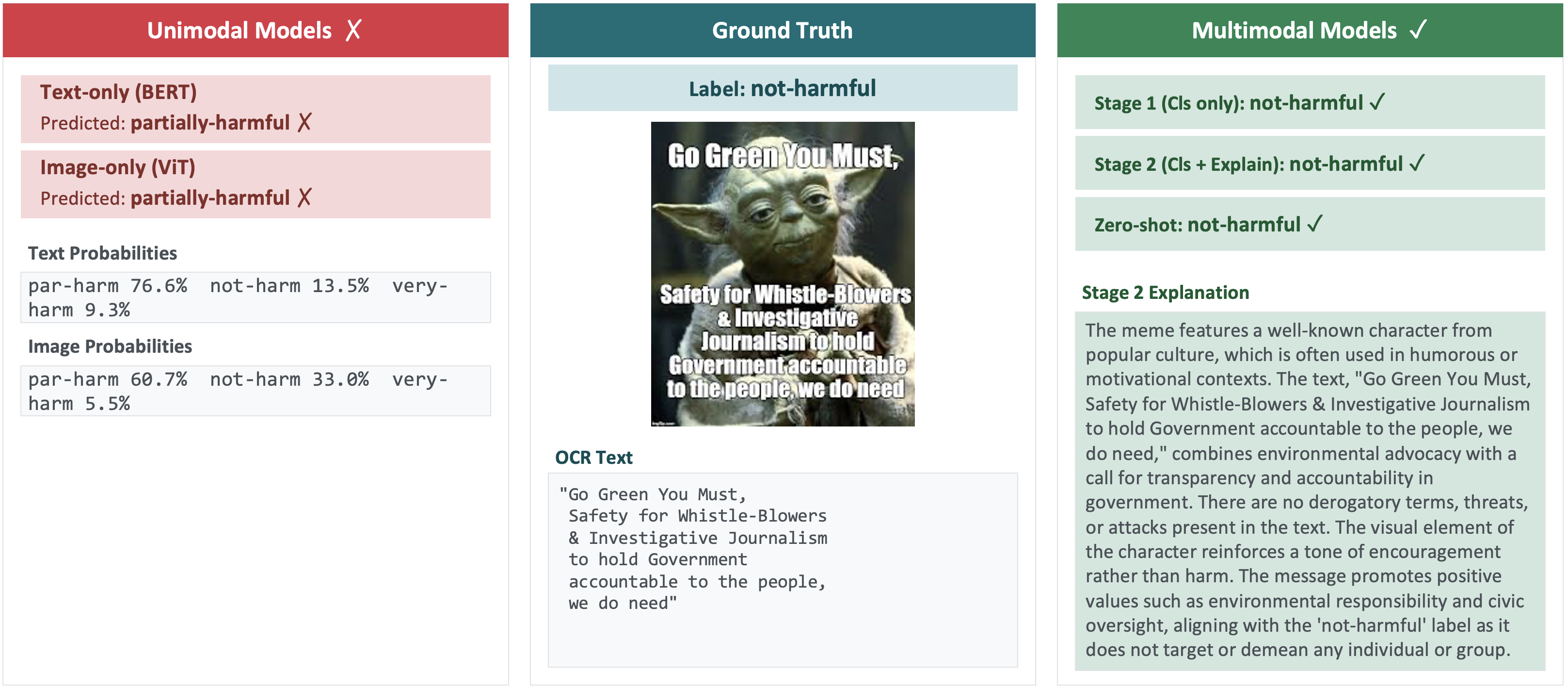}
\vspace{-0.2cm}
\caption{\textbf{Unimodal over-detection of harm.}
A benign meme misclassified as harmful by unimodal baselines due to lexical bias, while multimodal models correctly predict \texttt{not-harmful}. 
The \textit{Stage~II} rationale highlights supportive intent and lack of a target.}
\label{fig:qual-mm-not-harmful}
\vspace{-0.3cm}
\end{figure*}

\begin{figure*}[!tbh]
\centering
\includegraphics[width=0.86\textwidth]{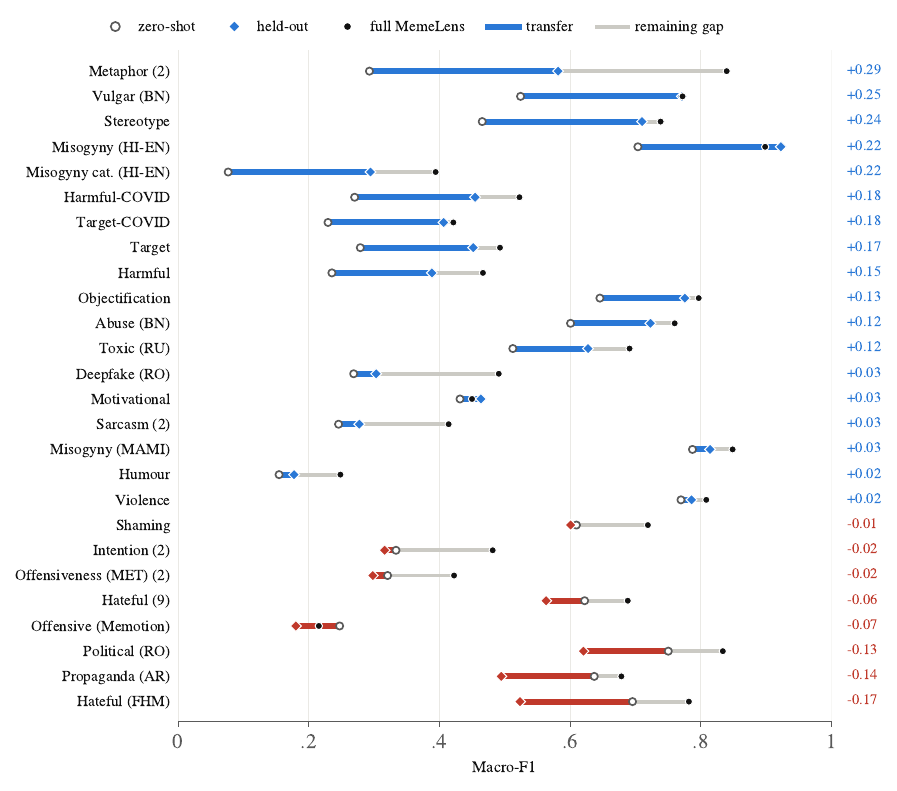}
\vspace{-0.2cm}
\caption{LOTO generalization by held-out task. Open circles denote ZS, diamonds HO, and black dots Full. Colored spans show the gain from ZS to HO, and gray spans the remaining gap to Full.}
\label{fig:loto_tasks}
\vspace{-0.35cm}
\end{figure*}

\paragraph{Unimodal baselines under-detect harm in highly harmful cases.} Figure~\ref{fig:qual-mm-very-harmful} presents a meme combining unverifiable case-count claims with the accusation ``\emph{hugest mass murderer ever}.'' Both unimodal baselines concentrate probability mass on \texttt{partially-harmful} (BERT: $83.6\%$, ViT: $60.0\%$), failing to capture the intensified and highly accusatory framing conveyed by the joint visual and textual context. In contrast, all multimodal systems we evaluated (Stage~I, Stage~II, and zero-shot MLLM) correctly predict \texttt{very-harmful}. The Stage~II explanation explicitly links this escalation to the co-occurrence of a political figure and stigmatizing claims, illustrating a failure mode that is not recoverable from either modality alone. More broadly, this example highlights how unimodal models underestimate harm when severity emerges from cross-modal reinforcement rather than isolated cues.

\paragraph{Unimodal baselines over-detect harm in benign cases.} Figure~\ref{fig:qual-mm-not-harmful} shows the complementary failure mode. A pop-culture character appears alongside advocacy text promoting whistle-blower protection and government accountability. Unimodal baselines again default to \texttt{partially-harmful} (BERT: $76.6\%$, ViT: $60.7\%$), over-relying on surface lexical cues such as ``\emph{government accountable}.'' All multimodal variants correctly predict \texttt{not-harmful}, with Stage~II explanations grounding the decision in the supportive tone, broader contextual framing, and absence of a targeted individual or group. This demonstrates how unimodal systems are prone to false positives when salient keywords appear without sufficient contextual grounding.

Overall, we saw that across many cases, Stage~I models struggle with metaphor, satire, and implicit references that require integrating information across modalities. We further saw that Stage~II training not only improves label accuracy, but also yields more faithful, interpretable rationales grounded in multimodal evidence. The consistent unimodal failures further demonstrate that harm detection in \textsc{MemeLens} is inherently multimodal and cannot be decomposed into independent text and image predictions. Instead, effective modeling requires capturing cross-modal dependencies and contextual nuance, highlighting the value of unified multimodal learning for robust meme understanding and future research.

\section{Leave-One-Out Generalization}
\label{appendix:loto_lolo}

This appendix provides detailed results for the leave-one-out experiments in Section~\ref{subsec:loto_lolo}. We train \textbf{Leave-One-Task-Out} (LOTO) models by excluding one task and its associated datasets, and \textbf{Leave-One-Language-Out} (LOLO) models by excluding one language. Each model is evaluated only on its corresponding held-out task or language. All models use the same Qwen3-VL-8B base model and training recipe as \memel{}. We compare the zero-shot model (\textbf{ZS}), the held-out model (\textbf{HO}), and full \memel{} (\textbf{Full}) on matched test instances.

\begin{figure}[t]
\centering
\includegraphics[width=\columnwidth]{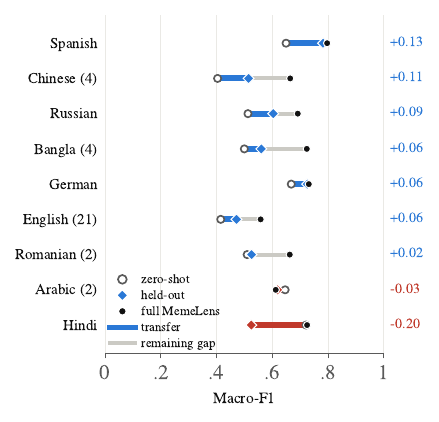}
\vspace{-0.2cm}
\caption{LOLO generalization by held-out language, using the same notation as Figure~\ref{fig:loto_tasks}.}
\label{fig:lolo_langs}
\vspace{-0.3cm}
\end{figure}

\begin{table}[!tbh]
\centering
\setlength{\tabcolsep}{3pt}
\scalebox{0.70}{%
\begin{tabular}{lcccc}
\hline
\textbf{Held-out task (dataset)} & \textbf{ZS} & \textbf{HO} & \textbf{Full} & \textbf{HO$-$ZS} \\
\hline
Misogyny-Cat. (MIMIC2024, HI-EN) & 13.3 & 65.5 & 76.6 & +52.2 \\
Target-COVID (HarMeme, EN) & 39.5 & 80.2 & 82.3 & +40.7 \\
Harmful-COVID (HarMeme, EN) & 30.1 & 68.4 & 74.8 & +38.3 \\
Harmful (HarMeme, EN) & 27.2 & 57.0 & 62.2 & +29.8 \\
Metaphor (MET-Meme, EN) & 28.8 & 56.8 & 86.7 & +27.9 \\
Toxic (Toxic-Memes, RU) & 54.2 & 78.9 & 86.6 & +24.7 \\
Stereotype (MAMI, EN) & 47.9 & 71.8 & 78.4 & +23.9 \\
Misogyny (MIMIC2024, HI-EN) & 70.3 & 92.3 & 89.9 & +21.9 \\
Metaphor (MET-Meme, ZH) & 42.4 & 62.6 & 86.6 & +20.2 \\
Abuse (BanglaAbuse, BN) & 59.4 & 77.9 & 78.7 & +18.5 \\
Objectification (MAMI, EN) & 64.4 & 79.4 & 83.5 & +15.0 \\
Sarcasm (BanglaAbuse, BN) & 43.4 & 58.4 & 67.4 & +14.9 \\
Target (HarMeme, EN) & 35.4 & 49.3 & 56.2 & +13.8 \\
Deepfake (RoMemes, RO) & 47.1 & 58.6 & 77.0 & +11.5 \\
Shaming (MAMI, EN) & 75.8 & 85.7 & 89.8 & +9.9 \\
Offensiveness (MET-Meme, ZH) & 54.1 & 63.4 & 83.0 & +9.3 \\
Vulgar (BanglaAbuse, BN) & 73.4 & 82.2 & 82.7 & +8.9 \\
Humour (Memotion, EN) & 19.4 & 27.2 & 35.2 & +7.8 \\
Hateful (Prop2Hate, AR) & 74.8 & 78.4 & 77.2 & +3.6 \\
Misogynous (MAMI, EN) & 78.6 & 81.5 & 84.9 & +2.9 \\
Hateful (MIMIC-Islamophobia, EN) & 66.0 & 68.7 & 70.7 & +2.7 \\
Intention (MET-Meme, ZH) & 40.4 & 43.0 & 71.0 & +2.6 \\
Violence (MAMI, EN) & 87.6 & 89.7 & 92.3 & +2.1 \\
Hateful (MUTE, BN) & 61.8 & 63.7 & 71.9 & +1.9 \\
Offensive (Memotion, EN) & 31.0 & 32.1 & 38.6 & +1.1 \\
Offensiveness (MET-Meme, EN) & 47.8 & 48.3 & 74.8 & +0.5 \\
Sarcasm (Memotion, EN) & 20.5 & 21.0 & 50.1 & +0.5 \\
Hateful (Multi3Hate, DE) & 67.2 & 67.2 & 75.4 & +0.0 \\
Propaganda (ArMeme, AR) & 76.1 & 74.9 & 78.9 & -1.2 \\
Hateful (MMHS, EN) & 42.7 & 40.9 & 61.4 & -1.8 \\
Motivational (Memotion, EN) & 65.3 & 63.3 & 63.7 & -2.0 \\
Hateful (Multi3Hate, ES) & 65.0 & 61.7 & 80.0 & -3.3 \\
Hateful (Multi3Hate, EN) & 74.1 & 69.0 & 74.1 & -5.2 \\
Political (RoMemes, RO) & 76.7 & 71.1 & 86.7 & -5.6 \\
Intention (MET-Meme, EN) & 43.3 & 37.1 & 52.4 & -6.2 \\
Hateful (Multi3Hate, ZH) & 55.7 & 49.2 & 77.0 & -6.6 \\
Hateful (FHM, EN) & 71.4 & 52.9 & 79.8 & -18.5 \\
Hateful (Multi3Hate, HI) & 72.1 & 49.2 & 75.4 & -23.0 \\
\hline
\end{tabular}
}
\vspace{-0.2cm}
\caption{Leave-One-Task-Out (LOTO): per held-out task, zero-shot (ZS), held-out model (HO), full \memel{} (Full), and transfer over zero-shot (HO$-$ZS). Accuracy (\%). Sorted by transfer.}
\label{tab:loto_full}
\vspace{-0.3cm}
\end{table}

\paragraph{Aggregate generalization.}
Both LOTO and LOLO substantially outperform the zero-shot model, achieving about $43\%$ of the accuracy gain obtained by full \memel{} training without using data from the held-out task or language. Accuracy gains are statistically significant in both settings. Macro-F1 also improves, with stronger evidence for LOLO than LOTO, suggesting that task-level generalization is more challenging for task-specific or rare labels.

\paragraph{Generalization across tasks and languages.} Figures~\ref{fig:loto_tasks} and~\ref{fig:lolo_langs} show considerable variation across held-out settings. Task-level gains are largest when related datasets provide supervision for similar label schemas, as observed for metaphor, fine-grained misogyny, stereotype, and the HarMeme harm and target tasks. In contrast, humour, sarcasm, and Memotion offensiveness show smaller gains. A few tasks, including FHM, Arabic propaganda, and Romanian political memes, perform better zero-shot than under held-out training.

The language-level results follow a similar pattern. Spanish, Chinese, and Russian benefit most from unified training, while Arabic changes little and Hindi declines. Overall, generalization gains tend to be larger when zero-shot performance is lower, indicating that unified training provides the greatest benefit where the base model is weaker.

These results show that \memel{} learns representations that generalize across tasks and languages even when the target task or language is excluded from training. However, the benefit depends on related supervision, and dataset-specific label definitions may still require in-domain training. This complements the single-dataset analysis in Section~\ref{subsec:diagnostic_single_dataset} and further supports unified training for cross-dataset generalization.

\begin{table}[t]
\centering
\setlength{\tabcolsep}{3pt}
\scalebox{0.70}{%
\begin{tabular}{lcccc}
\hline
\textbf{Held-out language (dataset)} & \textbf{ZS} & \textbf{HO} & \textbf{Full} & \textbf{HO$-$ZS} \\
\hline
Metaphor (MET-Meme, EN) & 28.8 & 70.8 & 86.7 & +42.0 \\
Harmful-COVID (HarMeme, EN) & 30.1 & 60.9 & 74.8 & +30.7 \\
Metaphor (MET-Meme, ZH) & 42.4 & 69.3 & 86.6 & +26.9 \\
Harmful (HarMeme, EN) & 27.2 & 52.1 & 62.2 & +24.9 \\
Toxic (Toxic-Memes, RU) & 54.2 & 78.9 & 86.6 & +24.8 \\
Offensiveness (MET-Meme, EN) & 47.8 & 71.4 & 74.8 & +23.5 \\
Offensiveness (MET-Meme, ZH) & 54.1 & 75.4 & 83.0 & +21.3 \\
Hateful (Multi3Hate, ZH) & 55.7 & 73.8 & 77.0 & +18.0 \\
Stereotype (MAMI, EN) & 47.9 & 65.4 & 78.4 & +17.5 \\
Sarcasm (BanglaAbuse, BN) & 43.4 & 59.5 & 67.4 & +16.1 \\
Hateful (Multi3Hate, ES) & 65.0 & 78.3 & 80.0 & +13.3 \\
Misogyny-Cat. (MIMIC2024, HI-EN) & 13.3 & 26.3 & 76.6 & +13.0 \\
Humour (Memotion, EN) & 19.4 & 32.1 & 35.2 & +12.7 \\
Intention (MET-Meme, ZH) & 40.4 & 52.3 & 71.0 & +11.9 \\
Objectification (MAMI, EN) & 64.4 & 73.6 & 83.5 & +9.1 \\
Sarcasm (Memotion, EN) & 20.5 & 29.6 & 50.1 & +9.1 \\
Deepfake (RoMemes, RO) & 47.1 & 55.2 & 77.0 & +8.0 \\
Abuse (BanglaAbuse, BN) & 59.4 & 67.0 & 78.7 & +7.6 \\
Offensive (Memotion, EN) & 31.0 & 38.7 & 38.6 & +7.6 \\
Hateful (Multi3Hate, DE) & 67.2 & 73.8 & 75.4 & +6.6 \\
Hateful (Prop2Hate, AR) & 74.8 & 80.2 & 77.2 & +5.4 \\
Misogyny (MIMIC2024, HI-EN) & 70.3 & 75.3 & 89.9 & +5.0 \\
Hateful (MIMIC-Islamophobia, EN) & 66.0 & 70.7 & 70.7 & +4.8 \\
Target (HarMeme, EN) & 35.4 & 39.5 & 56.2 & +4.0 \\
Vulgar (BanglaAbuse, BN) & 73.4 & 76.6 & 82.7 & +3.3 \\
Hateful (MUTE, BN) & 61.8 & 64.2 & 71.9 & +2.4 \\
Shaming (MAMI, EN) & 75.8 & 78.1 & 89.8 & +2.3 \\
Political (RoMemes, RO) & 76.7 & 78.9 & 86.7 & +2.2 \\
Intention (MET-Meme, EN) & 43.3 & 44.3 & 52.4 & +1.0 \\
Target-COVID (HarMeme, EN) & 39.5 & 40.4 & 82.3 & +0.9 \\
Motivational (Memotion, EN) & 65.3 & 65.1 & 63.7 & -0.1 \\
Hateful (MMHS, EN) & 42.7 & 42.0 & 61.4 & -0.7 \\
Propaganda (ArMeme, AR) & 76.1 & 74.9 & 78.9 & -1.2 \\
Hateful (Multi3Hate, EN) & 74.1 & 70.7 & 74.1 & -3.4 \\
Violence (MAMI, EN) & 87.6 & 82.7 & 92.3 & -4.9 \\
Misogynous (MAMI, EN) & 78.6 & 71.7 & 84.9 & -6.9 \\
Hateful (FHM, EN) & 71.4 & 55.7 & 79.8 & -15.6 \\
Hateful (Multi3Hate, HI) & 72.1 & 52.5 & 75.4 & -19.7 \\
\hline
\end{tabular}
}
\vspace{-0.2cm}
\caption{Leave-One-Language-Out (LOLO): per held-out language, zero-shot (ZS), held-out model (HO), full \memel{} (Full), and transfer over zero-shot (HO$-$ZS). Accuracy (\%). Sorted by transfer.}
\label{tab:lolo_full}
\vspace{-0.4cm}
\end{table}

\section{Data Release}
\label{apndix:release}
We made \memel{} dataset publicly available to the research community with the aim to enable further research in multilingual multimodal meme analysis. Our resource extends existing public datasets through label unification and the addition of explanations. Accordingly, we release our contribution under the CC BY-SA 4.0 license, that is, the Creative Commons Attribution-ShareAlike 4.0 International License: \url{https://creativecommons.org/licenses/by-sa/4.0/}. As the original datasets may be distributed under different licenses, we strongly encourage users of \memel{} to cite the source datasets and prior work where appropriate.

\end{document}